\documentclass[aps, pre, a4paper, floatfix, twocolumn, longbibliography, footinbib, amssymb]{revtex4-2}

\usepackage[colorlinks=true, hypertexnames=false]{hyperref}
\PassOptionsToPackage{linktocpage}{hyperref} %link numbers in TOC for arxiv
\hypersetup{
  colorlinks  = true, %Colour links instead of ugly boxes
  urlcolor = magenta!80!black,
  linkcolor  = blue!60!black, %Colour of internal links
  citecolor  = blue!60 %Colour of citations
}

\usepackage[T1]{fontenc}
\usepackage[utf8]{inputenc}
\usepackage{epsfig}
\usepackage[english]{babel}
\usepackage[table]{xcolor}
\usepackage{graphicx}
\usepackage{amsmath}
\usepackage{relsize}
\usepackage[percent]{overpic}
\usepackage{bm}
\usepackage{hyperref}
\usepackage{siunitx}
\usepackage{accents}
\usepackage{mathtools}
\usepackage{grffile}
\usepackage[normalem]{ulem}
\usepackage{pgf}
\usepackage{tikz}
\usepackage{times}

\usepackage[capitalize]{cleveref}

\usepackage{dsfont}  % \mathds{1} gives a double-stroke 1

\newcommand{\avg}[1]{{\left<#1\right>}}

\newcommand{\dd}{\mathrm{d}}
\newcommand{\ee}{\mathrm{e}}

\usepackage{mathtools}
\def\multiset#1#2{\ensuremath{\left(\kern-.3em\left(\genfrac{}{}{0pt}{}{#1}{#2}\right)\kern-.3em\right)}}

\usepackage{amsmath}

\newcommand{\A}{\bm{A}}

\newcommand{\bb}{\bm{b}}

\newcommand{\x}{\bm{x}}

\newcommand{\W}{\bm{W}}
\newcommand{\X}{\bm{X}}

\newcommand{\argmax}{\operatorname{arg\ max}}

\newcolumntype{P}[1]{>{\centering\arraybackslash}p{#1}}
\usepackage{algpseudocodex}
\usepackage{algorithm}

\algnewcommand\Input{\State\textbf{Input:} }
\algrenewcommand\Output{\State\textbf{Output:} }
\algnewcommand\Continue{\State\textbf{continue}}

\begin{document}

\title{Network reconstruction via the minimum description length principle}

\author{Tiago \surname{P. Peixoto}}
\email{tiago.peixoto@it-u.at}
\affiliation{Inverse Complexity Lab, IT:U Interdisciplinary Transformation University Austria, 4040 Linz, Austria}
\affiliation{Department of Network and Data Science, Central European University, Vienna, Austria}

\begin{abstract}
  A fundamental problem associated with the task of network reconstruction from
  dynamical or behavioral data consists in determining the most appropriate
  model complexity in a manner that prevents overfitting, and produces an
  inferred network with a statistically justifiable number of edges and their
  weight distribution. The \textit{status quo} in this context is based on
  $L_{1}$ regularization combined with cross-validation. However,
  besides its high computational cost, this commonplace approach unnecessarily
  ties the promotion of sparsity, i.e.\ abundance of zero weights, with weight
  ``shrinkage.'' This combination forces a trade-off between the bias introduced
  by shrinkage and the network sparsity, which often results in substantial
  overfitting even after cross-validation. In this work, we propose an alternative
  nonparametric regularization scheme based on hierarchical Bayesian inference
  and weight quantization, which does not rely on weight shrinkage to promote
  sparsity. Our approach follows the minimum description length (MDL) principle,
  and uncovers the weight distribution that allows for the most compression of
  the data, thus avoiding overfitting without requiring cross-validation. The
  latter property renders our approach substantially faster and simpler to
  employ, as it requires a single fit to the complete data, instead of many fits
  for multiple data splits and choice of regularization parameter. As a result,
  we have a principled and efficient inference scheme that can be used with a
  large variety of generative models, without requiring the number of
  reconstructed edges and their weight distribution to be known in advance. In a
  series of examples, we also demonstrate that our scheme yields systematically
  increased accuracy in the reconstruction of both artificial and empirical
  networks. We highlight the use of our method with the reconstruction of
  interaction networks between microbial communities from large-scale abundance
  samples involving in the order of $10^{4}$ to $10^{5}$ species, and
  demonstrate how the inferred model can be used to predict the outcome of
  potential interventions and tipping points in the system.
\end{abstract}

\maketitle

\section{Introduction}

Network reconstruction is the task of determining the unseen interactions
between elements of a complex system when only data on their behavior are
available---usually consisting of time-series or independent samples of their
states. This task is required when it is either too difficult, costly, or simply
impossible to determine the individual pairwise interactions via direct
measurements. This is a common situation when we want to determine the
associations between species only from their co-occurrence in population
samples~\cite{guseva_diversity_2022}, the financial market dependencies from the
fluctuation of stock prices~\cite{bury_statistical_2013}, protein structures
from observed amino-acid contact
maps~\cite{weigt_identification_2009,cocco_inverse_2018}, gene regulatory
networks from expression
data~\cite{dhaeseleer_genetic_2000,stolovitzky_dialogue_2007,vinciotti_consistency_2016,vinciotti_consistency_2016},
neural connectivity from fMRI, EEG, MEA, or calcium fluorescence imaging
data~\cite{maccione_experimental_2010,manning_topographic_2014,po_inferring_2024},
and epidemic contact tracing from
infections~\cite{braunstein_alfredo_network_2019}, among many other scenarios.

In this work we frame network reconstruction as a statistical inference
problem~\cite{peel_statistical_2022}, where we assume that the observed data has
been sampled from a generative statistical model that has a weighted network as
part of its set of parameters. In this case, the reconstruction task consists of
estimating these parameters from data. This generative framing contrasts with
other inferential approaches based on e.g. Granger
causality~\cite{sun_causal_2015} and transfer
entropy~\cite{orlandi_transfer_2014} in important ways: 1. It allows for a
principled framework that does not rely on \emph{ad hoc} and often arbitrary
confidence thresholds; 2. Can reliably distinguish between direct and indirect
interactions, to the extent that this information is obtainable from the data;
3. Enables uncertainty quantification and comparison between alternative
generative hypotheses; 4. Produces a generative model as a final result, which
can be used not only for interpretation, but also to predict the outcomes of
interventions and alternative conditions previously unseen in the data. These
same features make this type of approach preferable also to non-inferential
heuristics such as correlation
thresholding~\cite{bullmore_complex_2009,zogopoulos_approaches_2022}, known to
suffer from substantial
biases~\cite{cantwell_thresholding_2020,peel_statistical_2022}.

A central obstacle to the effective implementation of inferential network
reconstruction is the need for statistical regularization: A network with $N$
nodes can in principle have up to $O(N^{2})$ edges, and elementary maximum
likelihood approaches do not allow us to distinguish between the non-existence
of an edge and the existence of an edge with a low weight magnitude, resulting
in the inference of maximally dense networks, even when the true underlying
network is sparse, i.e. its number of edges is far smaller than the number of
possible edges. In this case, unregularized maximum likelihood would massively
overfit, incorporating a substantial amount of statistical noise in its
description, and thus would not meaningfully represent the true underlying
system. Becase of this, a robust inferential framework needs to include the
ability to determine the most appropriate model complexity from data, which in
this context means the right number of edges and their weight distribution, in a
manner that does not overfit.

Perhaps the most common strategy to avoid overfitting is $L_{1}$
regularization~\cite{hastie_statistical_2015}, where a penalty proportional to
the sum of the weight magnitudes is subtracted from the likelihood, multipiled
by a regularization parameter. This approach is a standard technique in
statistics and machine learning, perhaps most well known as the central
ingredient of LASSO regression~\cite{tibshirani_regression_1996}, and has been
used extensively as well for network reconstruction using multivariate Gaussian
models~\cite{meinshausen_high-dimensional_2006,yuan_model_2007,banerjee_model_2008,augugliaro_1-penalized_2020},
most famously as part of the GLASSO algorithm~\cite{friedman_sparse_2008}, but
also for the inverse Ising
model~\cite{bresler_reconstruction_2008,bresler_reconstruction_2010,
  ravikumar_high-dimensional_2010,bresler_efficiently_2015,nguyen_inverse_2017,
  vuffray_interaction_2016,lokhov_optimal_2018},
among other problem instances.

The $L_{1}$ regularization scheme has certain attractive properties: 1. It is
easy to implement; 2. It results in the existence of exact zeros for the
reconstructed weights; and 3. The $L_{1}$ penalty function is convex. When
combined with a likelihood function that is also convex (such as the one for the
multivariate Gaussian~\cite{friedman_sparse_2008} or the Ising
model~\cite{nguyen_inverse_2017}) the latter property means that this scheme
yields an overall convex objective function, thus enabling the use of very
efficient algorithms available only for this simpler class of optimization
problems. On the other hand, this scheme also has well-known significant
limitations~\cite{hastie_statistical_2015}: 1. The regularization parameter,
which determines the final sparsity of the reconstruction, needs to be provided
prior to inference, requiring the desired level of sparsity to be known
beforehand; 2. It causes ``shrinkage'' of the weights, i.e. the weights of the
non-zero edges also decrease in magnitude, introducing a bias. The first
disadvantage undermines this regularization approach in a central way, since it
requires as an \emph{input} what it should in fact provide as an \emph{output}:
the network sparsity. Because of this, the scheme is often used together with
cross-validation~\cite{friedman_sparse_2008, rothman_sparse_2008,
  liu_stability_2010} to determine the optimal value of the regularization
parameter. But not only does this significantly increase the computational cost
of the approach, but in the end also typically results in substantial
overfitting, as we demonstrate in this work.

Other regularization approaches also have been proposed for network
reconstruction, such as $\delta$-thresholding~\cite{aurell_inverse_2012},
decimation~\cite{decelle_pseudolikelihood_2014, franz_fast_2019},
EBIC~\cite{foygel_extended_2010}, adaptive LASSO~\cite{zou_adaptive_2006,
  meinshausen_relaxed_2007, beck_fast_2009}, MCP~\cite{zhang_nearly_2010},
SCAD~\cite{fan_variable_2001}, weight integration~\cite{wang_bayesian_2012},
alternative
priors~\cite{mohammadi_bayesian_2015,vogels_bayesian_2024,ni_bayesian_2022} with
fixed number of edges~\cite{dobra_bayesian_2011}, as well as
spike-and-slab~\cite{wang_scaling_2015,richard_li_bayesian_2019} and
horseshoe~\cite{carvalho_handling_2009,li_graphical_2019,li_using_2020} shapes.
These alternatives address some limitations of $L_{1}$, in particular its
detrimental shrinkage properties, but they all either still require cross
validation, rely on heuristics to determine the resulting sparsity, and/or make
strong assumptions on the shape of the weight distribution.

To the best of our knowledge, there is currently no regularization scheme that
is simultaneously: 1. Principled; 2. Effective at preventing overfitting; 2.
Algorithmically efficient; and 4. Nonparametric, in particular without requiring
the network sparsity to be known in advance.

In this work we fill this gap by developing a Bayesian regularization scheme
that follows the minimum description length (MDL)
principle~\cite{rissanen_information_2010,grunwald_minimum_2007}, and seeks to
find the optimal sparsity and weight distribution of the edges in a manner that
most compresses the observed data. Instead of weight shrinkage, our approach
relies on weight quantization as a means of quantifying the information
necessary to encode the inferred weights to a sufficient numerical precision,
and also makes no strong assumption on the shape of the weight distribution,
unlike $L_{1}$ and most Bayesian approaches previously proposed for this
problem~\cite{wang_bayesian_2012,mohammadi_bayesian_2015,vogels_bayesian_2024,dobra_bayesian_2011,wang_scaling_2015,richard_li_bayesian_2019,carvalho_handling_2009,li_graphical_2019,li_using_2020}.
The resulting approach does not require cross-validation and is also
algorithmically efficient, specially when combined with a recent algorithm that
can find the best potential edge candidates to be updated in subquadratic time
and in parallel~\cite{peixoto_scalable_2024}, making it applicable for the
reconstruction of networks with hundreds of thousands of nodes, or even more
depending on available computing resources. Combined with the fact that it does
not rely on any property of the generative model other than it being conditioned
on real-valued edge weights, our approach gives us a regularization scheme that
can be readily employed on a broad class of inferential reconstruction problems.

This paper is divided as follows. In Sec.~\ref{sec:framework} we describe our
inferential framework, and in Sec.~\ref{sec:l1} we discuss the problems with
$L_{1}$ regularization. In Sec.~\ref{sec:mdl} we present our alternative MDL
framework, and in Sec.~\ref{seq:algorithm} the inference algorithm we propose to
implement it, which we then evaluate with synthetic and empirical data. In
Sec.~\ref{sec:microbial} we conduct an empirical case study where we reconstruct
the large-scale network of interactions of microbial species from empirical data
and showcase how the inferred models can be used to predict the outcomes of
interventions, as well as to identify ``keystone'' species and tipping points.
We finalize in Sec.~\ref{sec:conclusion} with a conclusion.

\section{Inferential framework}\label{sec:framework}

The general scenario for inferential network reconstruction consists of some
data $\X$ that are assumed to originate from a generative model with a
likelihood
\begin{equation}
  P(\X | \W),
\end{equation}
where $\W \in \mathbb{R}^{N\times N}$ is a symmetric matrix corresponding to the
weights of an undirected graph of $N$ nodes. In general, we are not interested
on any particular \textit{a priori} constraint on $\bm W$, although we usually
expect it to be sparse, i.e.\ its number of non-zero entries scales as $O(N)$.
In many cases, the data are represented by a $N\times M$ matrix of $M$ samples,
with $X_{im}$ being a value associated with node $i$ for sample $m$, such that
\begin{equation}
  P(\X | \W) = \prod_{m=1}^{M}P(\bm x_{m} | \W),
\end{equation}
with $\bm x_{m}$ being the $m$-th column of $\X$. Alternatively, we may
have that the network generates a Markovian time series with likelihood
\begin{equation}
  P(\X | \W) = \prod_{m=1}^{M}P(\bm x_{m} | \bm x_{m-1},\W),
\end{equation}
given some initial state $\bm x_{0}$. Other variations are possible, but for our
purposes we need only to refer to a generic posterior distribution
\begin{equation}\label{eq:posterior}
  P(\W | \X) = \frac{P(\X| \W)P(\W)}{P(\X)}
\end{equation}
that we need to be able to compute up to a normalization constant. Our overall
approach is designed to work independently of what kind of generative model is
used, static or dynamic, with any type of likelihood, as long as it is
conditioned on edge weights according to the above equation. In
Appendix~\ref{app:models} we list some generative models that we will use as
examples in this work, but for now it is simpler if we consider the generative
model more abstractly.

Given Eq.~\ref{eq:posterior} we can proceed in a variety of ways, for example by
sampling from it, or computing expectations. In this work, we are interested in
the maximum \emph{a posteriori} (MAP) point estimate,
\begin{align}
  \widehat{\W} &= \underset{\W}{\argmax}\;P(\W | \X)\\
               &= \underset{\W}{\argmax}\;\log P(\X|\W) + \log P(\W),\label{eq:MAP}
\end{align}
corresponding to the most likely weighted network that generated the data. In
this setting, regularization is implemented by choosing a suitable prior
$P(\W)$, which should, ideally, properly penalize overly complex models.

In the following we will discuss $L_{1}$ regularization, and then present our
alternative approach based on the MDL principle.

\section{$L_{1}$ regularization, shrinkage, and overfitting}\label{sec:l1}

The most common choice to regularize Eq.~\ref{eq:MAP} are independent Laplace
distributions,
\begin{equation}\label{eq:laplace}
  P(\W|\lambda) = \prod_{i<j}\lambda \ee^{-\lambda|W_{ij}|}/2,
\end{equation}
which result in a $L_{1}$ penalty to the log-likelihood,
\begin{equation}
  \log P(\W|\lambda) = -\lambda \sum_{i<j}|W_{ij}| + {\textstyle {N\choose 2}}\log(\lambda/2).
\end{equation}
This choice offers some attractive properties. First, the function
$\log P(\W|\lambda)$ is convex, and therefore if $\log P(\X | \W)$ is also
convex with respect to $\bm W$ (which is true for key problem instances such as
the multivariate Gaussian or the Ising model), then Eq.~\ref{eq:MAP} amounts to
a convex optimization problem, which has a unique global solution with no other
local maxima, and can be solved efficiently~\cite{peixoto_scalable_2024}.
Second, this type of penalization promotes sparsity, i.e. as $\lambda$ is
increased, this will cause the values of $W_{{ij}}$ to converge in succession
exactly to zero at specific values of $\lambda$. This is in contrast to e.g.
$L_{2}$ regularization~\cite{tikhonov_stability_1943}, where an increase in the
penalty tends to merely reduce the magnitude of the weights continuously,
without making them exactly zero. In this way, the value of $\lambda$ amounts to
a proxy for the density of the inferred network.

However, this choice also suffers from important shortcomings. From a modeling
perspective, the prior of Eq.~\ref{eq:laplace} corresponds to a maximum entropy
distribution conditioned on the expected mean weight magnitude, including the
zeros. This amounts to a prior that is conditioned on previous knowledge about
the density of the network, encoded via the parameter $\lambda$. However, this
information is rarely known in advance, and is almost always one of the primary
objectives of inference in the first place. The likelihood of
Eq.~\ref{eq:laplace} is not helpful in this regard, since it is unbounded, and
diverges towards $\log P(\W|\lambda)\to\infty$ for $W_{ij}=0$ and
$\lambda\to \infty$, therefore we cannot infer $\lambda$ via MAP estimation.

The most common solution employed in practice is to select $\lambda$ via
$K$-fold cross-validation~\cite{friedman_sparse_2008, rothman_sparse_2008,
  liu_stability_2010}, i.e. by dividing the data matrix in $K$ disjoint subsets
$\X =\{\X_{1},\dots,\X_{K}\}$, and reconstructing the network with one subset
$k$ removed, i.e.
\begin{equation}\label{eq:kfold}
\widehat{\W}_{k}(\lambda) = \underset{\W}{\argmax}\;\log P(\X\setminus \X_{k}|\W) + \log P(\W|\lambda),
\end{equation}
and using the held-out data to compute the log-likelihood
$\mathcal{L}_{k}(\lambda) = \log P(\X_{k}|\widehat{\W}_{k}(\lambda))$.
Finally, $\lambda$ is chosen to maximize the mean held-out log-likelihood,
\begin{equation}
  \widehat{\lambda} = \underset{\lambda}{\argmax}\sum_{k}\mathcal{L}_{k}(\lambda).
\end{equation}
This standard technique is designed to prevent overfitting, since it relies on the
predictive performance of the inferred model $\widehat\W_{k}(\lambda)$ with
respect to the held-out data $\X_{k}$, which should deteriorate as soon as it
incorporates more statistical noise. Although widespread, this solution still
suffers from some key limitations. First, it increases the computational cost by
at least one or two orders of magnitude, since we need to perform the
optimization of Eq.~\ref{eq:kfold} $K$ times (a typical choice is $K=5$) and for
many values of $\lambda$ --- usually via a combination of grid and bisection
search. Furthermore, and even more importantly, this whole procedure does not
sufficiently reconcile the prevention of overfitting with the promotion of
sparsity, as we now demonstrate.

\begin{figure}
  \setlength{\tabcolsep}{0pt}
  \begin{tabular}{cc}
    \multicolumn{2}{c}{\begin{overpic}[width=\columnwidth]{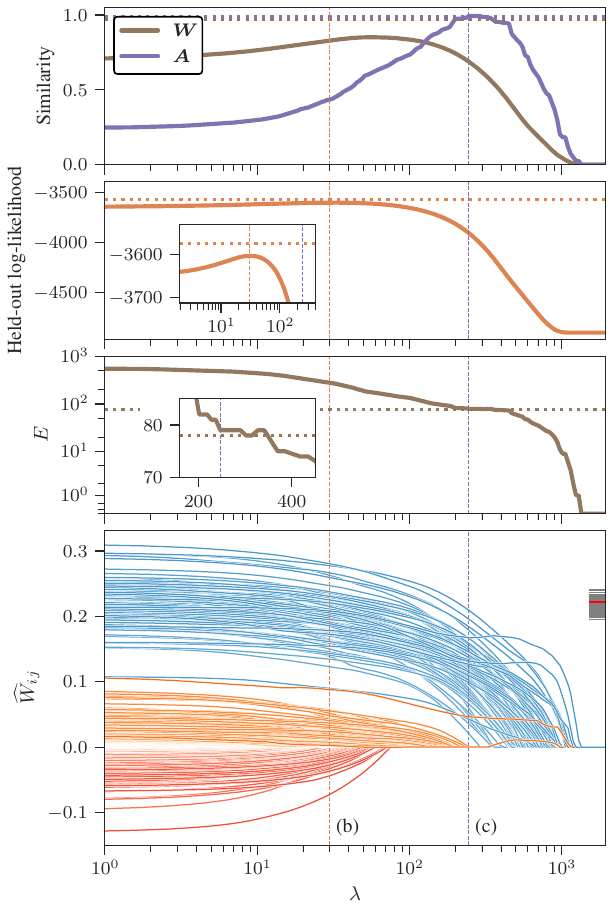}\put(0,98){(a)}\end{overpic}}\\
    (b) $\;\lambda = 30.8$ & (c) $\;\lambda = 246.6$ \\
    \begin{overpic}[width=.45\columnwidth]{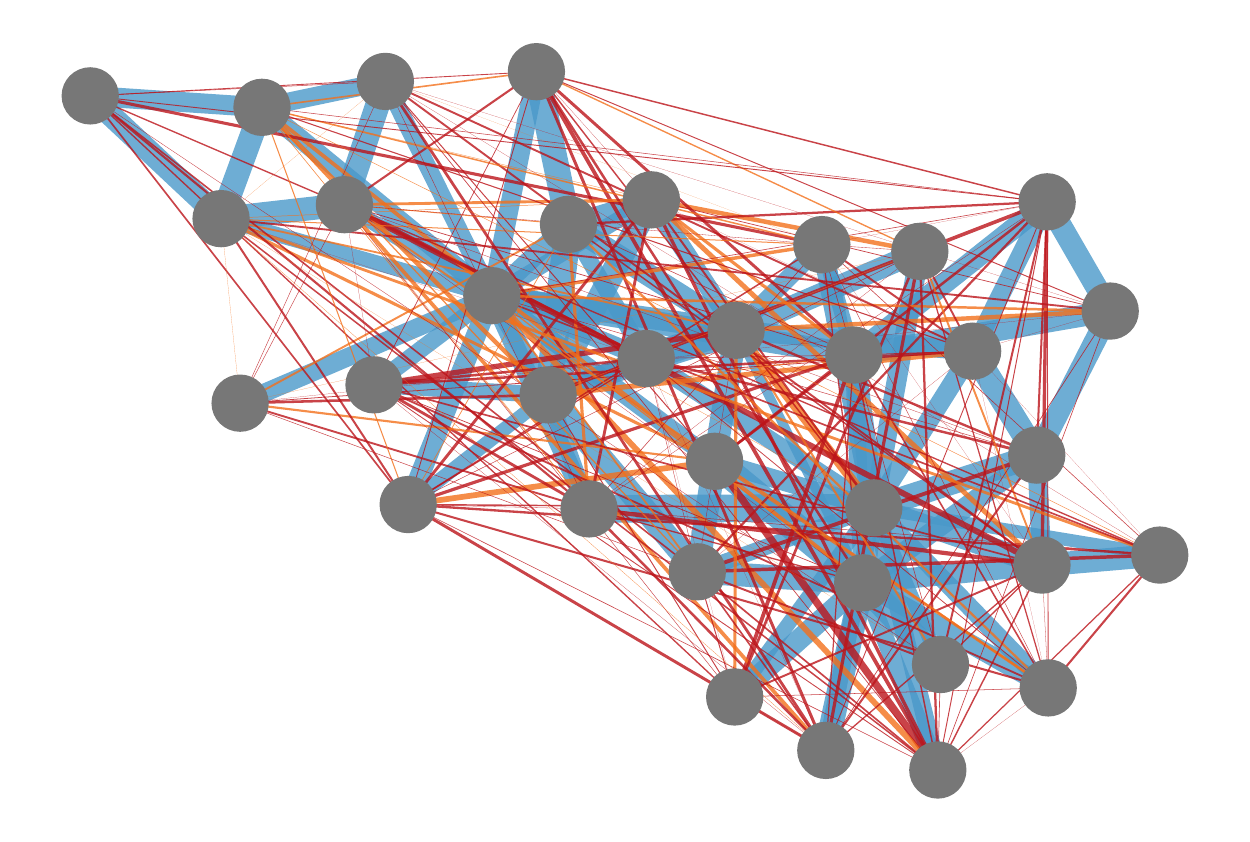}\end{overpic} &
    \begin{overpic}[width=.45\columnwidth]{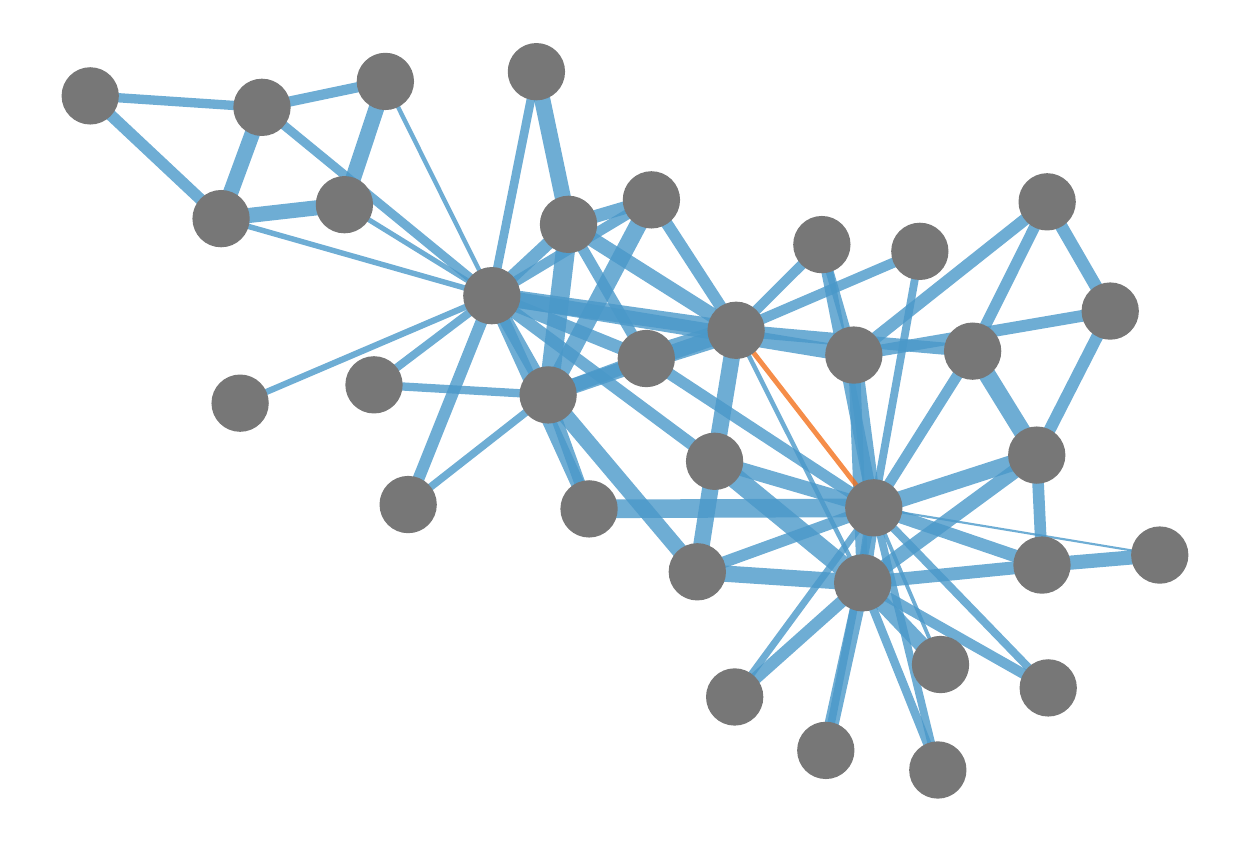}\end{overpic}
  \end{tabular}
  \caption{\textbf{$L_{1}$ regularization overfits when combined with cross
      validation}. This example considers the reconstruction of the weighted
    Karate club network~\cite{zachary_information_1977} ($N=34$ nodes and $E=78$
    edges, weight values sampled i.i.d. from a normal distribution with mean
    $0.22$ and standard deviation $0.01$), based on $M=1,000$ transitions from
    the kinetic Ising model with a random initial state, using Eqs.~\ref{eq:MAP}
    and~\ref{eq:laplace}, for a range of values of the regularization strength
    $\lambda$. Panel (a) shows, from top to bottom, the Jaccard similarity
    between inferred and true weights and binarized edges, the mean held-out
    likelihood for a 5-fold cross validation, the number of inferred non-zero
    edges, and the individual values of inferred weights (with true nonzero
    edges shown in blue, and true zero-valued entries shown in red). The grey
    horizontal lines at the right margin of the bottom of panel (a) show the
    true weight values. The vertical dashed lines mark the values of $\lambda$
    that maximize (b) the mean held-out likelihood and (c) the binarized Jaccard
    similarity. The inferred network for these two values of $\lambda$ are shown
    in panels (b) and (c), respectively, with edge weights represented as
    thickness and the color representing whether it is a true (blue) or spurious
    (red) edge. In panel (a), all dashed horizontal lines, as well as the red
    line at the right margin of the bottom panel, mark the results obtained with
    the MDL regularization of Sec~\ref{sec:mdl}.\label{fig:shrinkage}}
\end{figure}

For our analysis it will be useful to evaluate the accuracy of a reconstruction
$\widehat \W$ via the Jaccard similarity $s(\W,\widehat\W)\in [0,1]$,
\begin{equation}
  s(\W,\widehat\W) = 1-\frac{\sum_{i< j} |W_{ij}-\widehat W_{ij}|}{\sum_{i< j} |W_{ij}|+|\widehat W_{ij}|},
\end{equation}
and likewise for the binarized adjacency matrix, $A_{ij}(\W)=\{1 \text{ if } |W_{ij}|>0, \text{ else } 0\}$, with
$s(\A(\W),\A(\widehat\W))$.

In Fig.~\ref{fig:shrinkage} we consider the reconstruction of Zachary's karate
club network~\cite{zachary_information_1977}, with $N=34$ nodes and $E=78$ edges
serving as the nonzero couplings of a kinetic Ising model (see
Appendix~\ref{app:models}), with each corresponding value of $\W$ sampled from a
normal distribution with mean $N/2E = 0.22$ and standard deviation $0.01$. After
sampling a single time series of $M=1,000$ transitions given a random initial
state, we performed the reconstruction for a range of values of $\lambda$,
including also a 5-fold cross validation procedure for each value. As seen in
Fig.~\ref{fig:shrinkage}a, the increase in $\lambda$ simultaneously sparsifies
and shrinks the magnitude of the inferred edge weights. The maximum of the mean
held-out log-likelihood obtained from cross validation is achieved for a
value $\lambda\approx 30.8$, for a network with $E=278$ nonzero weights, shown
in Fig.~\ref{fig:shrinkage}b. The weighted and unweighted similarities with the
true network are $0.83$ and $0.44$, respectively, affected primarily by the
abundance of spurious edges incorporated into the reconstruction. If we increase
instead to $\lambda\approx 246.6$, the similarities become $0.67$ and $0.99$,
respectively, with $E=79$---much closer to the true value---illustrating the
inherent trade-off between weight magnitude preservation and sparsity. Although
for this value of $\lambda$ the binarized network is much more accurate, the
average weight value is only $0.11$---around half of the true value---which
significantly degrades the model's predictive performance. (We observe that
without knowing the true network, it would not have been possible to determine
the value of $\lambda$ that yields the largest unweighted similarity, since
neither the model likelihood nor the cross-validation procedure offer any
indication in this regard.)

It is clear from this example that the $L_{1}$ regularization obtained with the
Laplace prior forces an unnecessary dichotomy between sparsity and low weight
magnitudes. The detrimental bias of $L_{1}$ to obtain regression coefficients is
well known~\cite{fan_variable_2001} and alternative such as adaptive
LASSO~\cite{zou_adaptive_2006, meinshausen_relaxed_2007, beck_fast_2009},
MCP~\cite{zhang_nearly_2010}, and SCAD~\cite{fan_variable_2001} exist which
mitigate this problem; but they all involve \emph{ad hoc} parameters that need
to be determined via cross-validation, without the algorithmic attractiveness of
$L_{1}$, since these alternatives are no longer convex. As a result, these
alternatives are substantially more cumbersome to use.

Ideally, we should be able to determine which edges are better modeled by
setting them to zero, without inherently reducing the magnitude of the non-zero
ones. This is an inference problem on its own, and instead of committing
beforehand to which kind of weight distribution is more appropriate, a more
robust approach consists in formulating an open-ended class of distributions,
and learning its precise shape \emph{a posteriori} from the data, following the
principle of maximum parsimony. This is precisely what we achieve in the next
section.

\section{Weight quantization and the minimum description length principle}\label{sec:mdl}

Our purpose is to implement a principled regularization scheme that does not
rely on weight shrinkage, promotes sparsity, and obviates the need for
cross-validation. Ideally, we would like a choice of prior $P(\W)$ that achieves
this directly via the MAP estimation
\begin{align}\label{eq:objective}
  \widehat{\W} = \underset{\W}{\argmax}\;\log P(\X|\W) + \log P(\W).
\end{align}
More precisely, we want to evoke the inherent connection between Bayesian
inference and data compression by interpreting the joint likelihood $P(\X, \W)$
in terms of the description length~\cite{rissanen_information_2010,grunwald_minimum_2007}
\begin{align}
  \Sigma(\X,\W) &= -\log P(\X, \W) \\
                &= -\log P(\X|\W) -\log P(\W),
\end{align}
where the first term in the right hand side of the last equation corresponds to
the amount of information required (e.g. in bits if base 2 is used for the
logarithm) to encode the data $\X$ when the parameters $\W$ are known, and the
second term corresponds to the amount of information required to encode the
parameters $\W$. In this setting, the prior for $\W$ should act as a penalty
counteracting the likelihood term that prevents overly complex models from being
inferred. The result of the optimization of Eq.~\ref{eq:objective} would then
amount to the most compressive model---i.e. with the shortest description
length---striking the ideal balance between quality of fit and model complexity,
extirpating as much as possible statistical noise from the model description.

However, the above equivalence with compression is not valid when the weights
$\bm W$ are modeled as continuous values with infinite precision. In this case,
their prior $P(\bm W)$ is necessarily a probability \emph{density} function, and
hence $-\log P(\bm W)$ cannot be interpreted as information
content~\footnote{Erroneously interpreting the log of a probability density
  $P(\bm W)$ as information could not only lead to meaningless negative values
  when $P(\bm W) > 1$, but also its value would change \emph{arbitrarily} under
  an exchange of variables $\W'=\bm U(\W)$, with $\bm U(\bm z)$ bijective,
  leading to a new probability density
  $P'(\W') = P(\bm U^{-1}(\W'))\left|\det [\nabla_{\bm z}\bm U^{-1}(\bm z)|_{\bm z = \W'}]\right|$,
  and a new value $-\log P'(\bm W')$ arbitrarily smaller or larger than
  $\log P(\bm W)$, even though it refers to a fully equivalent model.}. This is
exactly the reason why we cannot determine the value of $\lambda$ of the Laplace
prior of Eq.~\ref{eq:laplace} via MAP estimation: the most likely value is
\emph{not} necessarily the most compressive one. In order to properly quantify
the model complexity, we need be to be explicit about the precision with which
we describe the parameters $\bm W$, which means that they must be discretely
distributed according to a probability \emph{mass} function.

With the objective of precisely quantifying the model complexity, we proceed in
several steps. First, we account for the sparsity of the model by introducing an
auxiliary variable $\A$ corresponding to the binary adjacency matrix that
determines the values of $\W$ that are going to be non-zero. At first, we assume
$\A$ is sampled from a uniform distribution conditioned on the total number of
edges $E$,
\begin{equation}\label{eq:er}
  P(\A|E) = \frac{\delta_{\sum_{i<j}A_{ij},E}}{\textstyle {{N \choose 2} \choose E}},
\end{equation}
and the value of $E$ is itself sampled from a uniform prior
$P(E) = 1/[{N\choose 2}+1]$---ultimately an unimportant constant added simply
for completeness. This choice will allow us to benefit from sparsity, as it will
lead to shorter description lengths for graphs with fewer edges. Given $\bm A$,
we then sample the weights $\bm W$ as
\begin{equation}
  P(\bm W | \bm A) = \prod_{i<j}P(W_{ij})^{A_{ij}}\delta_{W_{ij},0}^{1-A_{ij}},
\end{equation}
where $P(W_{ij})$ is the probability of a non-zero weight $W_{ij}$, which is the
same for all $(i,j)$. Importantly, in order to account for precision,
$P(W_{ij})$ must be a probability mass function with a support over a discrete
set of values. In particular, we may, for example, consider the weights to be
the outcome of a quantization procedure
\begin{equation}
  W_{ij} = \Delta\left\lceil\frac{Y_{ij}}{\Delta}\right\rfloor,
\end{equation}
where $Y_{ij}$ is some continuous auxiliary value, $\lceil\cdot\rfloor$ is the
round operation, and $\Delta$ represents the precision considered, such that we
have a conditional probability mass
$P(W_{ij}|Y_{ij},\Delta) = \delta_{W_{ij},\Delta\left\lceil Y_{ij}/\Delta\right\rfloor}$.
Note that we do not assume that the true values of $W_{ij}$ are actually sampled
in this way; instead this prior simply articulates the exact precision $\Delta$
with which we are performing the inference.

We can then proceed by choosing a continuous distribution for the auxiliary
values $\bm Y$, e.g. a Laplace distribution
\begin{equation}
  P(Y_{ij}|\lambda) = \lambda \ee^{-\lambda|Y_{ij}|}/2,
\end{equation}
in which case we obtain a probability mass
\begin{multline}\label{eq:qlap}
  P(W_{ij}|\lambda,\Delta)
  = \frac{\int P(W_{ij}|Y_{ij},\Delta)P(Y_{ij}|\lambda)\,\dd Y_{ij}}{1-\int P(0|Y_{ij},\Delta)P(Y_{ij}|\lambda)\,\dd Y_{ij}} \\
  \quad=\begin{cases}
      \ee^{-\lambda|W_{ij}|}(\ee^{\lambda\Delta} - 1)/2, &\text{ if }\; \begin{aligned}
                                                                          W_{ij}&=\Delta\left\lceil W_{ij}/\Delta\right\rfloor \\
                                                                          W_{ij}&\neq 0
                                                                        \end{aligned}\\
      0, &\text{ otherwise, }
   \end{cases}
 \end{multline}
 where we have excluded the value $W_{ij}=0$ to be consistent with our
 parametrization based on $\bm A$. Stopping at this point we would have solved
 two issues: 1. Via quantization we can now properly evaluate the description
 length of the weights; 2. We can benefit from sparsity in a manner that is
 independent from the weight magnitudes. However, by choosing the Laplace prior,
 we remain subject to weight shrinkage, since the reduction of the description
 length is still conditioned on their overall magnitudes---besides the
 discretization, Eq.~\ref{eq:qlap} still amounts to $L_{1}$ regularization.
 Therefore, our final step is to add one more level to our Bayesian hierarchy,
 and try to learn the weight distribution from the data, instead of assuming
 that the probability decays according to the Laplace distribution. For that, we
 assume that the discretized weights are sampled directly from an arbitrary
 discrete distribution
\begin{equation}
  P(W_{ij}|\bm p, \bm z) = \sum_{k=1}^{K}\delta_{W_{ij},z_{k}}p_{k},
\end{equation}
where $\bm z=\{z_{k}\}$ is a set of $K$ real values representing discrete weight
categories, and $\bm p = \{p_{k}\}$ are their corresponding probabilities, with
$\sum_{k}p_{k}=1$. From this we obtain
\begin{equation}
  P(\bm W | \bm p, \bm z, \bm A) = \prod_{k}p_{k}^{m_{k}} \times \prod_{i<j}\delta_{W_{ij},0}^{1-A_{ij}},
\end{equation}
with $m_{k}=\sum_{i<j}A_{ij}\delta_{W_{ij},z_{k}}$ being the counts of weight
category $k$, and with a marginal distribution
\begin{align}
  P(\W|\bm z,\bm A)
  &= \int P(\W|\bm p, \bm z, \bm A)P(\bm p)\,\dd \bm p \\
  &= \frac{\prod_{k}m_{k}!}{E!} \times {K+E-1 \choose E}^{-1}\times \prod_{i<j}\delta_{W_{ij},0}^{1-A_{ij}},
\end{align}
assuming a uniform prior density $P(\bm p) = (K-1)!$ over the $K$ simplex. The
latter distribution allows for a combinatorial interpretation: it corresponds to
a uniform distribution among the $E$ edges of the weights with fixed counts
$\bm m = \{m_{k}\}$, with probability
$P(\bm W | \bm m) = \frac{\prod_{k}m_{k}!}{E!} \prod_{i<j}\delta_{W_{ij},0}^{1-A_{ij}}$,
and the uniform distribution of the counts $\bm m$ themselves, with probability
$P(\bm m | E)={K+E-1 \choose E}^{-1}$. This interpretation allows us to fix a
minor inconvenience of this model, namely that it allows for a weight category
$z_{k}$ to end up empty with $m_{k}=0$, which would require us to consider
category labels that never occur. Instead, we can forbid this explicitly by
using a uniform prior $P(\bm m | E)={E-1 \choose K-1}^{-1}$ over strictly
non-zero counts $\bm m$, leading to a slightly modified model
\begin{equation}\label{eq:pW}
  P(\W|\bm z,\bm A)
  = \frac{\prod_{k}m_{k}!}{E!} \times {E-1 \choose K-1}^{-1}\times \prod_{i<j}\delta_{W_{ij},0}^{1-A_{ij}}.
\end{equation}
Finally, we need a probability mass for the weight categories $\bm z$
themselves. At this point, however, the importance of this choice is minor,
since we expect $K \ll E$. A simple choice is to evoke the quantized Laplace
once more, with
\begin{multline}\label{eq:z}
  P(z_{k}|\lambda,\Delta)\\
  \quad=\begin{cases}
      \ee^{-\lambda|z_{k}|}(\ee^{\lambda\Delta} - 1)/2, &\text{ if }\; \begin{aligned}
                                                                          z_{k}&=\Delta\left\lceil z_{k}/\Delta\right\rfloor \\
                                                                          z_{k}&\neq 0
                                                                        \end{aligned}\\
      0, &\text{ otherwise, }
   \end{cases}
 \end{multline}
 with $P(\bm z|\lambda,\Delta,K) = \prod_{k=1}^{K}P(z_{k}|\lambda,\Delta)$.
 Although this choice implies that there will be a residual amount of shrinkage
 with respect to the weight categories, it will not dominate the regularization,
 unless the weight categories would otherwise diverge---something that can
 happen with certain kinds of graphical models such as the multivariate Gaussian
 when pairs of nodes have exactly the same values in $\bm X$. Because of this,
 we found this residual shrinkage to be beneficial in such corner cases, and
 otherwise we do not observe it to have a noticeable effect~\footnote{It is in
   fact easy to consider other choices for $P(\bm z)$ that eliminate residual
   shrinkage. A simple one is to sample the weight categories uniformly from the
   set of real numbers that are representable with $q$ bits (e.g. using the IEEE
   754 standard for floating-point arithmetic~\cite{noauthor_ieee_2008}),
   leading to $P(\bm z)= {2^{q} - 1\choose K}^{-1}\approx 1/2^{qK}$. This gives
   virtually indistinguishable results to the choice of Eq.~\ref{eq:z} in most
   examples we investigated, except, as mentioned in the main text, in
   situations where the weight categories would tend to diverge with
   $|z_{k}|\to\infty$, where we found Eq.~\ref{eq:z} to yield a more convenient
   model as its residual shrinkage prevents this from happening in these edge
   cases.}.

Lastly, we need a
prior over the number of weight categories $K$, which we assume to be uniform in the allowed range,
with $P(K|\A)=(1-\delta_{E,0})/E + \delta_{E,0}\delta_{K,0}$.
Putting it all together we have,
\begin{align}
  P(\W|\lambda,\Delta)
         &= \sum_{\bm z, K, \bm A,E}P(\W|\bm z, \bm A)P(\bm z|\lambda,\Delta,K)\times \nonumber\\[-1em]
         &\qquad\qquad\qquad\qquad P(K|\bm A)P(\A|E)P(E)\\
         &=\frac{\prod_{k}m_{k}!\times \ee^{-\lambda \sum_k |z_k|}(\ee^{\lambda\Delta} - 1)^{K}}{E!{E-1 \choose K-1}2^{K}\max(E,1){\textstyle {{N \choose 2} \choose E}}[{N\choose 2}+1]},\label{eq:W_prior}
\end{align}
where the remaining quantities $\bm m$, $\bm z$, $K$, $\bm A$, and $E$ in
Eq.~\ref{eq:W_prior} should be interpreted as being functions of $\bm W$. Note
that this model recovers the one using Eq.~\ref{eq:qlap} when $K=E$, i.e. every
weight belongs to its own category, up to a unimportant factor
$1/(E+1)$---although we expect to obtain significantly smaller description
length values with Eq.~\ref{eq:W_prior}, since it can exploit regularities in
the weight distribution and determine the most appropriate precision of the
weights. We emphasize that, differently from Eq.~\ref{eq:qlap}, in
Eq.~\ref{eq:W_prior} the parameter $\Delta$ controls the precision of the weight
categories $\bm z$, not directly of the weights themselves. The precision of the
weights $\W$ are primarily represented by the actual finite categories $\bm z$.

The model above is conditioned on two hyperparameters, $\lambda$ and $\Delta$,
which in principle can be both optimized, assuming a sufficiently uniform prior
over them (therefore, contributing to a negligible constant to the description
length). In practice, the effect of such an optimization is marginal, and it is
more convenient to select a small value for $\Delta$, such as
$\Delta = 10^{-8}$, which allows us to treat the weight categories as continuous
for algorithmic purposes. For $\lambda$ we use an initial value of $\lambda=1$,
and optimize if necessary.

We observe that we have $-\log P(\bm z|\lambda,\Delta) = O(K\Delta)$ and
$-\log P(\W|\bm z, \bm A) = O(E\log E - E\log K)$, and therefore we will
typically have $K\ll E$ when the description length is minimized, in which case
the prior for $\bm z$ will have only a minor contribution to the overall
description length. Therefore, our approach is not very sensitive to alternative
choices for the weight category distribution, $P(\bm z|\lambda,\Delta)$. We
observe also that the dominating term $-\log P(\W|\bm z,\bm A)$ from
Eq.~\ref{eq:pW} is completely invariant to bijective transformations of the
weights, provided the categories $\bm z$ are transformed in the same manner.
Therefore, any scale or shape dependence is driven solely by the relatively
minor contribution of $\log P(\bm z|\lambda,\Delta)$, proportional only to the
number of weight categories. Since the scale parameter $\lambda$ can be
optimized, this means that our approach does not impose any constraints on the
scale and only very weak constraints on the shape of the weights (see also
footnote~\cite{Note2}).

Differently from the Laplace prior of Eq.~\ref{eq:laplace}, Eq.~\ref{eq:W_prior}
implements a nonparametric regularization of the weights---not because it lacks
parameters, but because the shape of the distribution and number of parameters
adapt to what can be statistically justified from the data. In particular, the
number and position of the weight categories, as well as the number of edges,
will grow or shrink depending on whether they can be used to compress the data,
and provide a sufficiently parsimonious network reconstruction, without
overfitting.

The prior of Eq.~\ref{eq:W_prior} incorporates the properties we initially
desired, i.e. 1. It penalizes model complexity according to the description
length, 2. It exploits and promotes sparsity, and 3. It does not rely on weight
shrinkage. However, these improvements come at a cost: The prior is no longer a
convex function on $\bm W$. Because of this, this approach requires special
algorithmic considerations, which we address in the following.

\section{Inference algorithm}\label{seq:algorithm}

Whenever Eq.~\ref{eq:objective} corresponds to a convex optimization objective,
it can be performed with relatively straighforward algorithms. The simplest one
is coordinate descent~\cite{wright_coordinate_2015}, in which every entry in the
$\bm W$ matrix is optimized in sequence, while keeping the others fixed. This
yields an algorithm with a complexity of $O(\tau N^2)$, provided the
optimization of each entry can be done in time $O(1)$ (e.g. via bisection
search), and $\tau$ is the number of iterations necessary for sufficient
convergence. A recent significant improvement over this baseline was proposed in
Ref.~\cite{peixoto_scalable_2024}, where an algorithm was presented that solves
the same optimization problem in subquadratic time, typically with a
$O(N\log^{2}N)$ complexity, although the precise scaling will depend on the
details of the data in general. This algorithm works as a greedy version of
coordinate descent, where a subset of the entries are repeatedly proposed to be
updated according to the NNDescent algorithm~\cite{dong_efficient_2011}. The
latter implements a stochastic search that involves keeping a list of candidate
edges incident on each node in the network as an endpoint, and updating this
list by investigating the list of second neighbors in the candidate network.
This algorithm typically converges in log-linear time, allowing the overall
reconstruction algorithm also to converge in subquadratic time, in this way
converging orders of magnitude faster than the quadratic baseline of the
coordinate descent algorithm.

Although these algorithms cannot be used unmodified to our problem at hand, due
its non-convexity and the presence of additional latent variables, we can use
them as building blocks. We will consider different types of updates, according
to the different types of latent variable we are considering, given our initial
$\W$ (initially empty with $W_{ij}=0$) and our set of weight categories $\bm z$
(also empty initially), which are applied in arbitrary order, until the
description length no longer improves:
\begin{description}
  \item[Edge updates] We use the algorithm of Ref.~\cite{peixoto_scalable_2024}
        to find the $\kappa N$ best zero-valued entries of $\W$ as update
        candidates, usually with $\kappa=1$, according to the pairwise ranking
        score
        \begin{equation}
          \pi_{ij} = \max\left[\ln P(\W^{+}|X),\; \ln P(\W^{-}|X)\right],
        \end{equation}
        where $\W^{+}$ is the new matrix $\W$ with entry $W_{ij}$ replaced by
        the smallest positive weight category in $\bm z$, and likewise with
        the smallest magnitude negative value for $\W^{-}$. If $\bm z$ is
        currently empty, we replace the score by the gradient magnitude,
        \begin{equation}
          \pi_{ij} = \left|\frac{\partial\ln P(\W'|X)}{\partial W_{ij}'}\right|_{\W'=\W}.
        \end{equation}
        The algorithm of Ref.~\cite{peixoto_scalable_2024} can obtain this set
        in typical time $O(N\log^{2}N)$. We call $\mathcal{E}$ the union of this
        set with the currently nonzero values of $\W$. We then visit every entry
        $(i,j)$ in $\mathcal{E}$, and try to perform an update
        $W_{ij}\to W_{ij}'$, accepting only if it improves the
        objective function, chosen uniformly at random from the options:
        \begin{enumerate}
          \item $W_{ij}'\in \bm z$ chosen according to a random bisection
                search~\footnote{Random bisection search will eventually succeed
                in finding the global minimum of a multimodal objective, hence
                it should in general be preferred over deterministic bisection, unless
                the objective function is convex.} maximizing the objective
                function, with each midpoint sampled uniformly at a random.
          \item $W_{ij}'\in \mathbb{R}$ and $W_{ij}'\notin \bm z$ chosen
                according to a random bisection search in the interval of weights
                allowed for the problem. This will introduce a new weight
                category with $\bm z' = \bm z \cup \{W_{ij}'\}$.
          \item $W_{ij}' \leftarrow 0$, if $W_{ij} > 0$.
        \end{enumerate}
  \item[Edge moves] Given the same set $\mathcal{E}$ obtained as described
        above, we iterate over the nodes $i$ of the network with nonzero degree,
        and for an incident edge $(i,j)$ chosen uniformly at random, and an
        entry $(i,u)$ chosen uniformly at random from the subset of
        $\mathcal{E}$ with $W_{iu}=0$, we swap their values, i.e.
        $W_{iu}\leftarrow W_{ij}$ and $W_{ij} \leftarrow 0$, and accept if this improves the
        objective function.
  \item[Edge swaps] We chose uniformly at random two nonzero entries in $\bm W$,
        $(i,j)$ and $(u,v)$ involving four distinct nodes, and we swap their
        endpoints, i.e. $(W_{iv},W_{ij})\leftarrow (W_{ij}, W_{iv})$ and
        $(W_{uj},W_{uv})\leftarrow (W_{uv}, W_{uj})$, and accept only in case of improvement.
  \item[Weight category values] We iterate over each weight category
        $z_{k} \in \bm z$ and perform an update $z_{k}\leftarrow z_{k}'$, with $z_{k}'$
        being the result of a bisection search over the allowed values.
  \item[Weight category distribution] We consider the distribution of weight
        categories across the nonzero entries of $\W$ as a clustering problem,
        using the merge-split algorithm described in
        Ref.~\cite{peixoto_merge-split_2020}, composed of the following moves,
        which are accepted if it improves the quality function:
        \begin{enumerate}
          \item \emph{Merge.} Two categories $z_{k}$ and $z_{l}$ are removed and
          merged as a single new category $z_{m}$, with a value chosen
          with a bisection search optimizing the objective function. The
          number of categories reduces by one.
          \item \emph{Split.} A single category $z_{m}$ is split in two new
          categories $z_{k}$ and $z_{l}$, chosen initially uniformly
          random in the range $[z_{m-1},z_{m+1}]$. The entries are
          repeatedly moved between these categories if it improves the
          quality function, and the category values $z_{k}$ and $z_{l}$
          are updated according to bisection search, until convergence.
          The number of categories increases by one.
          \item \emph{Merge-split.} The two steps above are combined in
          succession, so that the entries are redistributed in two
          potentially new categories, and while their total number remains
          the same.
        \end{enumerate}
\end{description}

The workhorse of this overall procedure is the algorithm of
Ref.~\cite{peixoto_scalable_2024} which reduces the search for candidate updates
to a subquadratic time. The remaining steps of the algorithm operate on this set
with linear complexity, and therefore a single ``sweep'' of all latent variables
can be accomplished under the same algorithmic complexity as
Ref.~\cite{peixoto_scalable_2024}, since it dominates. In practice, due to the
extra latent variables and optimization steps, this algorithm finishes in time
around 5 to 10 times longer than when using $L_{1}$ regularization for a single
value of $\lambda$, but is typically much faster than doing cross-validation and
scanning over many values of $\lambda$. A reference C++ implementation of this
algorithm is freely available as part of the \texttt{graph-tool} Python
library~\cite{peixoto_graph-tool_2014}.

\subsection{Assessment of the algorithm with synthetic data}\label{seq:evaluation}

\begin{figure}
  \setlength{\tabcolsep}{0pt}
  \begin{tabular}{ccc}
    &High-school friendships~\cite{moody_peer_2001} & American football~\cite{girvan_community_2002} \\
    &($N=291$, $E=1,396$) & ($N=115$, $E=613$)\\
    \rotatebox[origin=l]{90}{\hspace{3em}Kinetic Ising model} &
    \begin{overpic}[width=.45\columnwidth]{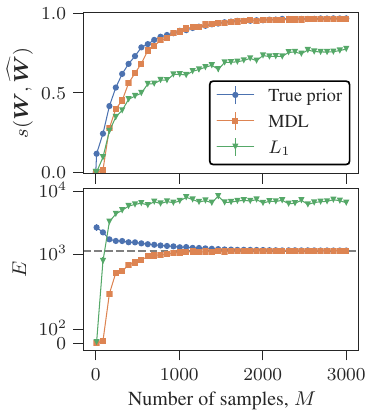}\put(-5,100){(a)}\end{overpic} &
    \begin{overpic}[width=.45\columnwidth]{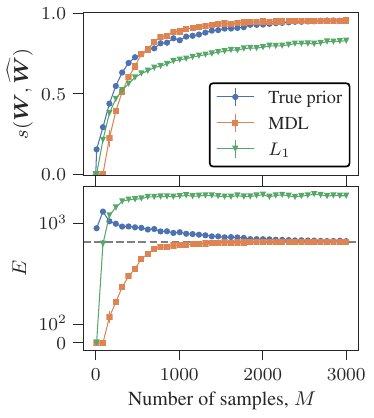}\put(0,100){(b)}\end{overpic} \\
    \rotatebox[origin=l]{90}{\hspace{2em}Equilibrium Ising model} &
    \begin{overpic}[width=.45\columnwidth]{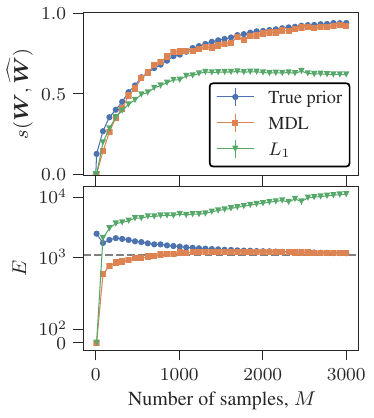}\put(-5,100){(c)}\end{overpic} &
    \begin{overpic}[width=.45\columnwidth]{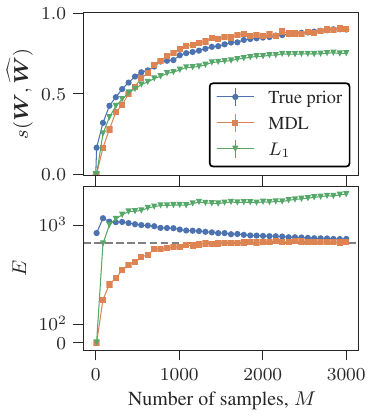}\put(0,100){(d)}\end{overpic}
  \end{tabular}
  \caption{Inference results for the kinetic and equilibrium Ising model for two
    empirical networks, as indicated in the legend, with true weights sampled as
    described in the text, using both the true prior, our MDL regularization, as
    wel as $L_{1}$ regularization with 5-fold cross-validation. The individual
    panels show the Jaccard similarity $s(\W,\hat\W)$ between the inferred and
    true networks, as well as the number $E$ of inferred non-zero edges (the
    dashed horizontal line shows the true value).\label{fig:synthetic}}
\end{figure}

In Fig.~\ref{fig:synthetic} we demonstrate the use of our algorithm on synthetic
data generated from empirical networks. We simulate the kinetic and equilibrium
Ising models~\footnote{To sample from the equilibrium Ising model we use MCMC
  with the Metropolis-Hastings criterion~\cite{metropolis_equation_1953}. To
  determine equilibration we computed the $\hat{R}$ of parallel
  chains~\cite{vehtari_rank-normalization_2021}, as well as the effective sample
  size (ESS).} (see Appendix~\ref{app:models}) with true weights sampled i.i.d.
from a normal distribution with mean $\mu=1/\avg{k}$, with $\avg{k}=2E/N$, and
standard deviation $\sigma=0.01$. We compare our approach with a prior
corresponding to the true distribution, i.e.
\begin{equation}
  P(\bm W |\mu,\sigma,\bm A) = \prod_{i<j}\left[\frac{\ee^{-(W_{ij}-\mu)^{2}/2\sigma^{2}}}{\sqrt{2\pi}\sigma}\right]^{A_{ij}}\delta_{W_{ij},0}^{1-A_{ij}}.
\end{equation}
We observe that our MDL regularization performs very closely to the true prior,
except for very few data, in which case the MDL tends to produce an empty
network, whereas the true prior in fact overfits and yields a network with a
number of edges larger than the true value. As we already explained, MAP
estimation does not guarantee proper regularization when the priors correspond
to probability densities---even if it happens to be the correct one. The overall
tendency of the MDL approach is, by design, to err towards simpler models,
instead of more complex ones, when the data is insufficient to provide a good
accuracy.

As already discussed previously in Sec.~\ref{sec:l1}, $L_{1}$ regularization
with cross-validation performs substantially worse than our MDL approach,
overfitting the data with networks that are orders of magnitude denser than they
should be. Unlike using the true prior or MDL, for the examples of
Fig.~\ref{fig:synthetic} the results obtained with $L_{1}$ show no sign of
converging asymptotically to true network as the data increases with
$M\to\infty$. This behavior further highlights the general unsuitability of the
$L_{1}$ approach even in situations where the data is abundant. We should also
mention that the results obtained with $L_{1}$ took typically one to two orders
of magnitude longer to be computed, due to the need to optimize over the
parameter $\lambda$ using a bisection search.

% We also compare in Fig.~\ref{fig:shrinkage} the result of the MDL regularization
% with $L_{1}$ for the Karate Club example, where we obtain a performance
% with MDL that is superior to $L_{1}$ for any value of $\lambda$.

\subsection{Joint reconstruction with community detection}

The use of the auxiliary sparse network $\A$ and its uniform prior $P(\A|E)$ given
by Eq.~\ref{eq:er} opens the opportunity for more structured priors that can
take advantage of inferred structure for further data compression, and in this way
provide improved regularization~\cite{peixoto_network_2019}. One generically
applicable approach is to use the degree-corrected stochastic block model
(DC-SBM)~\cite{karrer_stochastic_2011}, here in its microcanonical
formulation~\cite{peixoto_nonparametric_2017}, with a likelihood
\begin{equation}
  P(\A |\bb,\bm k,\bm e) = \frac{\prod_{r<s}e_{rs}!\prod_{r}e_{rr}!!\prod_{i}k_{i}!}{\prod_{i<j}A_{ij}\prod_{i}A_{ii}!!\prod_{r}e_{r}!},
\end{equation}
where $\bb=\{b_{i}\}$ is the node partition, with $b_{i}\in\{1,\dots,B\}$ being
the group membership of node $i$, $\bm k =\{k_{i}\}$ is the degree sequence,
with $k_{i}$ being the degree of node $i$, and $\bm e = \{e_{rs}\}$ is the group
affinity matrix, with $e_{rs}$ being the number of edges between groups $r$ and
$s$, or twice that if $r=s$. With this model, we can formulate the problem of
reconstruction according to the joint posterior
\begin{equation}
  P(\W, \A, \bb | \X) = \frac{P(\X|\W)P(\W|\A)P(\A|\bb)P(\bb)}{P(\X)},
\end{equation}
where the marginal distribution
$P(\A|\bb) = \sum_{\bm k, \bm e}P(\A |\bb,\bm k,\bm e)P(\bm k|\bm e)P(\bm e)$ is
computed using the priors described in Ref.~\cite{peixoto_nonparametric_2017},
in particular those corresponding to the hierarchical (or nested)
SBM~\cite{peixoto_hierarchical_2014}.

As discussed in Ref.~\cite{peixoto_network_2019}, the use of this kind of
structured prior can improve regularization, since it provides more
opportunities to reduce the description length, whenever the underlying network
is not completely random. The obtained community structure is often also useful
for downstream analysis and interpretation. Algorithmically, this extension of
our method is straightforward, as we need only to include merge-split partition
moves~\cite{peixoto_merge-split_2020}, as done to infer the SBM.

We refer the reader to Ref.~\cite{peixoto_nonparametric_2017} for more details
about the SBM formulation, and to Ref.~\cite{peixoto_network_2019} for its
effect in network reconstruction.

\subsection{Node parameters}

Besides the weighted adjacency matrix $\W$, many generative models posses
additional parameters, typically values $\bm\theta \in \mathbb{R}^{N}$ with
$\theta_{i}$ being a value associated with node $i$ (e.g. a local field in the
Ising model, or the mean and variance for a multivariate Gaussian). Although we
could in principle include them in the diagonal of $\bm W$, these values often
have a different interpretation and distribution, or there are more than one of
them per node, and hence they need to be considered separately.

In principle, since the size of $\bm\theta$ is a constant when performing
network reconstruction, regularizing their inference is a relatively smaller
concern. However, it would be incongruous to simply neglect this aspect in our
approach. Luckily, it is a simple matter to adapt our quantized prior for
$\bm W$ into one for $\bm\theta$, the only difference being that we no longer
need auxiliary variables to handle sparsity, and we can allow zeros in the
discrete categories $u_{k}$ for $\bm\theta$, sampled in this case as
\begin{multline}\label{eq:u}
  P(u_{k}|\lambda_{\theta},\Delta_{\theta})\\
  \quad=\begin{cases}
    \ee^{-\lambda_{\theta}|u_{k}|}\sinh(\lambda_{\theta}\Delta_{\theta}), &\text{ if }\; \begin{aligned}
                                                                      u_{k}&=\Delta_{\theta}\left\lceil u_{k}/\Delta_{\theta}\right\rfloor \\
                                                                      u_{k}&\neq 0,
                                                                        \end{aligned}\\
    1-\ee^{-\lambda_{\theta}\Delta_{\theta}}, &\text{ if } u_{k}=0,\\
    0, &\text{ otherwise. }
   \end{cases}
\end{multline}
Proceeding as before gives us
 \begin{multline}
  P(\bm\theta|\lambda_{\theta},\Delta_{\theta})
  =\prod_{k}n_{k}!\times \ee^{-\lambda \sum_k |u_k|}\\\times \frac{\sinh(\lambda_{\theta}\Delta_{\theta})^{K_{\theta}-\mathds{1}_{0\in\bm u}}(1-\ee^{-\lambda_{\theta}\Delta_{\theta}})^{\mathds{1}_{0\in\bm u}}}{N!{N-1 \choose K_{\theta}-1}N},\label{eq:theta_prior}
\end{multline}
where  $n_{k}=\sum_{i}\delta_{\theta_{i},u_{k}}$ are the category counts, $K_{\theta}$ is the number of  categories.

\subsection{Empirical evaluation, contrast to $L_{1}$ regularization and decimation}

Now we perform a comparative evaluation of our MDL regularization for empirical
data where true network is not available. We will consider the votes of all
$N=623$ deputies in the lower house of the Brazilian
congress~\cite{peixoto_network_2019}, during the 2007 to 2011 term, as variables
$x_{i}\in \{-1,0,1\}$, corresponding to ``no'', ``abstain'', and ``yes'',
respectively. This period corresponds to $M=619$ voting sessions, where all deputies
cast a vote on a piece of legislation. We use a modified version of the
equilibrium Ising model that allows for zero values (see
Appendix~\ref{app:models}) to capture the latent coordination between individual
deputies.

\begin{figure}
  \setlength{\tabcolsep}{0pt}
  \begin{tabular}{cc}
    \begin{overpic}[width=.5\columnwidth,trim=0cm 0cm 1cm 1cm]{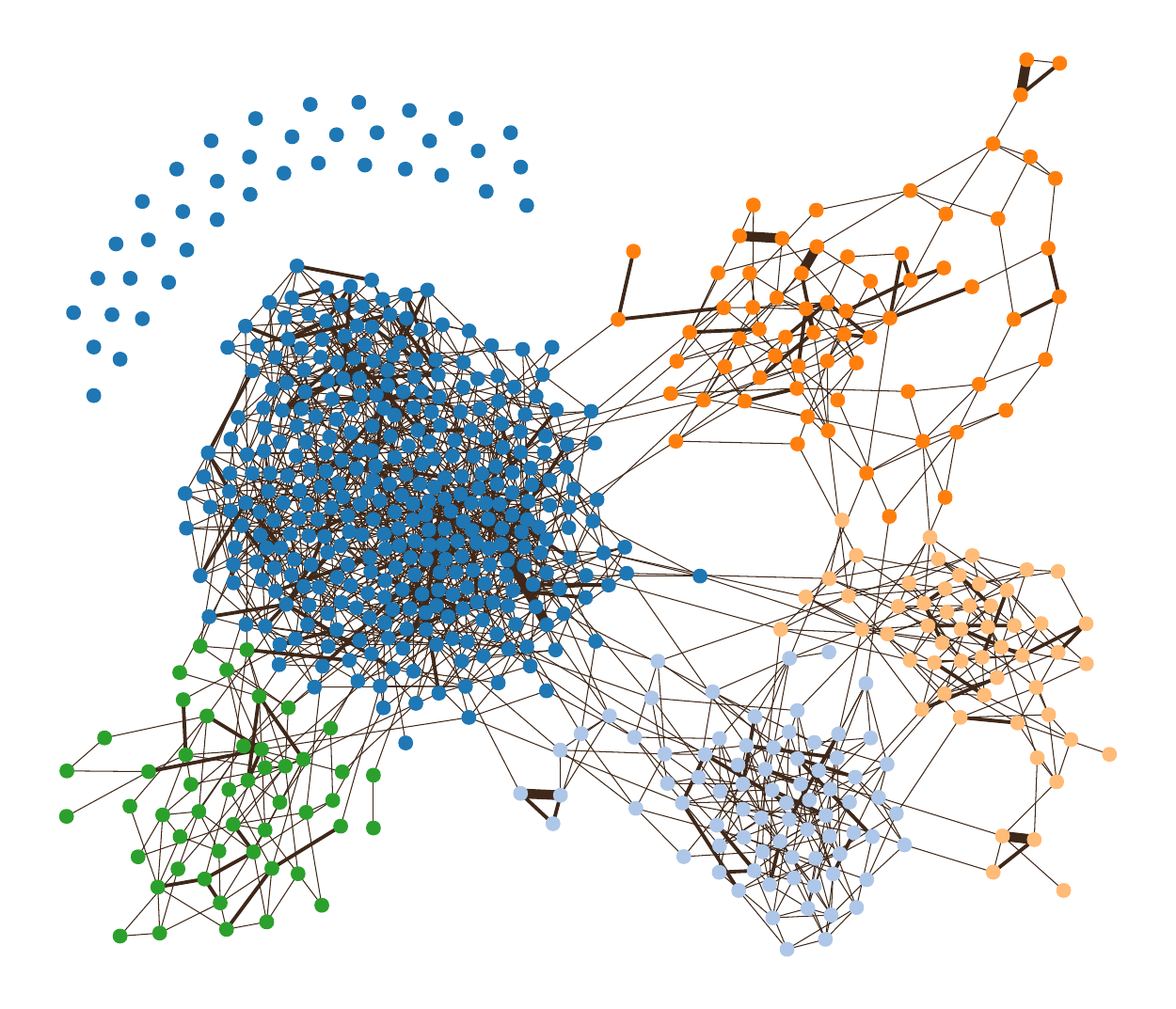}\put(0,75){(a)}\end{overpic} &
    \begin{overpic}[width=.5\columnwidth,trim=0cm 0cm 1cm 1cm]{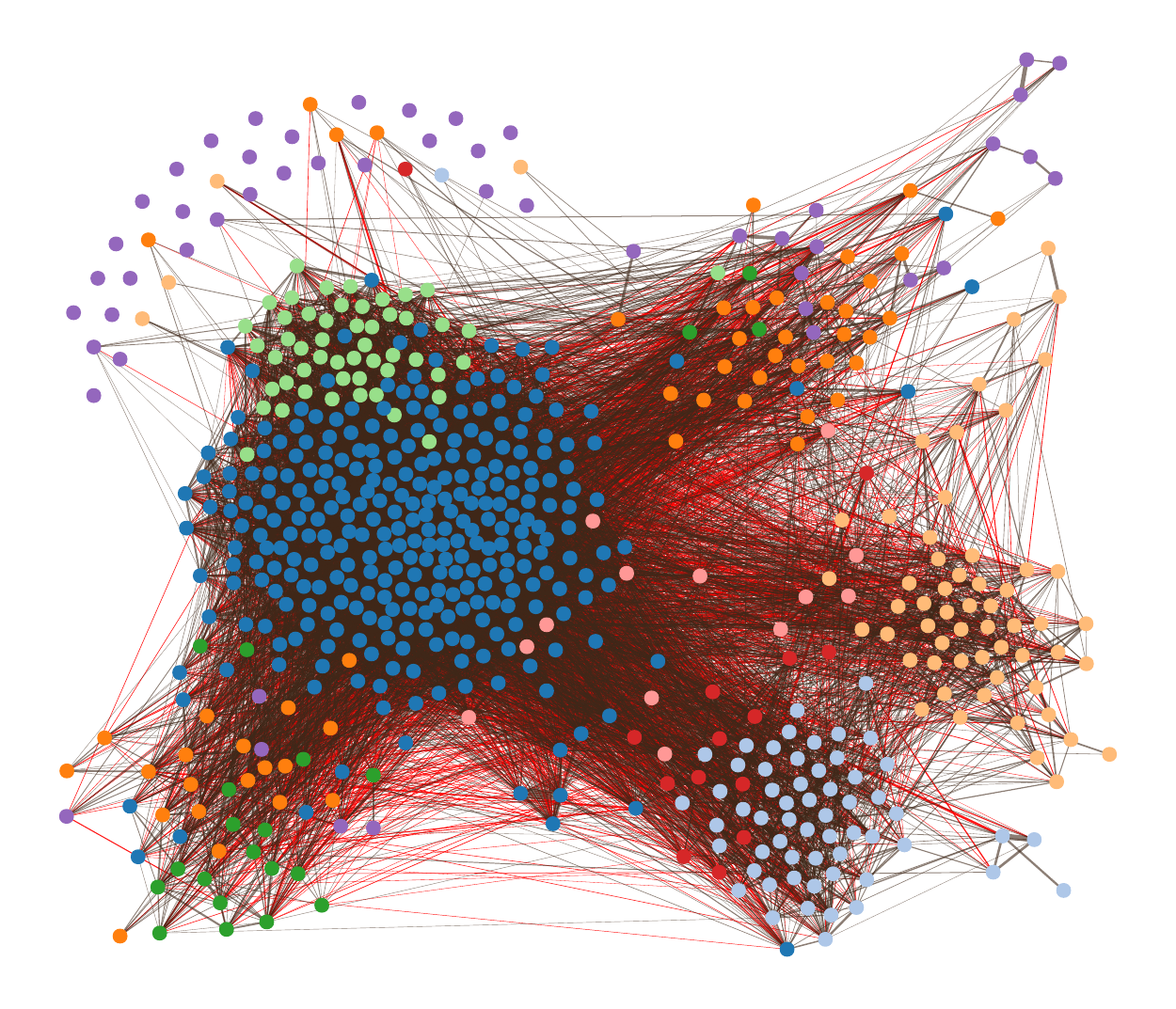}\put(0,75){(b)}\end{overpic}
  \end{tabular}
  \begin{tabular}{cc}
    \begin{overpic}[width=.72\columnwidth]{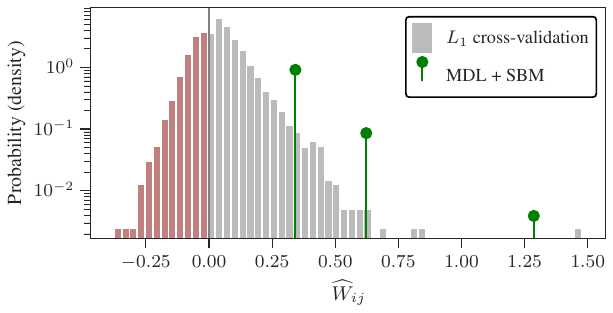}\put(0,52){(c)}\end{overpic} &
    \begin{overpic}[width=.28\columnwidth]{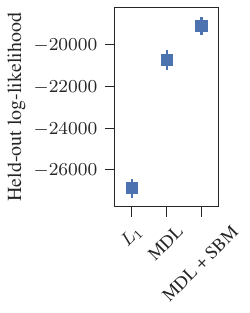}\put(0,105){(d)}\end{overpic}
  \end{tabular}
  \caption{Reconstructed networks of interactions between $N=623$ members of the
    lower house of the Brazilian congress~\cite{peixoto_network_2019}, during
    the 2007 to 2011 term, corresponding to $M=619$ voting sessions. Panel (a)
    shows the network inferred with MDL regularization, with edge weights
    corresponding to their thickness, and the node colors indicating the division
    found with the SBM incorporated into the regularization. Panel (b) shows the
    result with $L_{1}$ together with 5-fold cross validation, and negative
    weights shown in red. The colors indicate the group assignments found by
    fitting an SBM to the resulting network. Panel (c) shows the weight
    distributions obtained with both methods, and (d) the mean held-out
    likelihood of a 5-fold cross validation with each method, including also the
    MDL version without the SBM.\label{fig:camara}}
\end{figure}

In Fig.~\ref{fig:camara} we show the result of the reconstruction using both our
MDL scheme and $L_{1}$ regularization together with 5-fold cross-validation.
Similarly to the Karate Club example of Fig.~\ref{fig:shrinkage}, the $L_{1}$
inference incorporates one order of magnitude more non-zero edges ($E=10,911$)
when compared with the MDL method ($E=1,540$). The edges obtained with $L_{1}$
are mostly of low magnitude (see Fig.~\ref{fig:camara}c), and based on our
analysis for synthetic data, are much more likely to be overfitting. We can find
additional evidence of this by performing cross-validation also with the MDL
method, as shown in Fig.~\ref{fig:camara}d, which yields a significantly
improved predictive performance of the held-out data. Both the predictive
performance and description length improve when performing MDL regularization
together with community detection, in accordance with
Ref.~\cite{peixoto_network_2019}.

The interpretation of the two inferred models is fairly different. Although they
share similar large-scale structures (as shown by the node colors indicating the
inferred communities), the one inferred by $L_{1}$ possesses weights that are
often negative, indicating some amount of direct antagonism between deputies,
i.e. conditioning on the ``yes'' vote of one deputy will increase the
probability of a ``no'' vote of another. Instead, the model inferred by MDL
possesses only positive weights, divided in three categories, indicating an
overwhelming tendency towards selective agreement, instead of direct antagonism.
The groups it finds also align very well with the available metadata: the
largest group corresponds to the governing coalition, whereas the remaining are
opposition parties forming fragmented blocks~\footnote{Both methods also lead to
  inferences of isolated nodes. These correspond to deputies who are mostly
  absent from voting sessions, and more typically abstain.}.

\begin{figure}
  \includegraphics[width=\columnwidth]{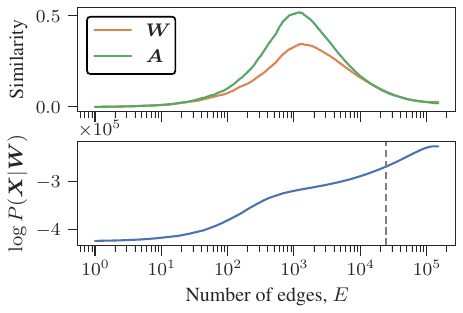}
  \caption{Decimation procedure~\cite{decelle_pseudolikelihood_2014} employed on
    the same data as Fig.~\ref{fig:camara}. The bottom panel shows the maximum
    likelihood growing monotonically as a function of the number of nonzero
    edges considered during decimation, and the top shows the similarity
    (weighted and binarized) with the network inferred via MDL. The vertical
    dashed line corresponds to the value $E=24,196$, obtained using the stopping
    criterion proposed in
    Ref.~\cite{decelle_pseudolikelihood_2014}.\label{fig:decimation}}
\end{figure}

We take the opportunity to compare with another regularization scheme called
``decimation,'' proposed in Ref.~\cite{decelle_pseudolikelihood_2014}. That
approach is not based on shrinkage, and instead proceeds by first performing an
unregularized maximum likelihood, resulting in a full network with no zero
weights, then forcing a small number of the weights with smallest magnitude to
zero, and proceeding recursively on the remaining entries in the same manner,
until sufficiently many entries have been set to zero. Besides the increased
computational cost of inferring the weights multiple times, and starting with a
full network---thus imposing an overall complexity of at least $O(N^{2})$---this
approach offers no obvious, principled criterion to determine when to stop the
decimation. In Ref.~\cite{decelle_pseudolikelihood_2014} a heuristic stopping
criterion was proposed by observing a point at which the likelihood no longer
changes significantly in synthetic examples sampled from the true model.
However, there is no guarantee that such a signal will exist in empirical data.
In Fig.~\ref{fig:decimation} we show for our Brazilian congress example the
model likelihood offers no clear indication of when to stop the decimation and
how many edges should be inferred in the end. The Jaccard similarity with the
network inferred via MDL remains low through the whole procedure, peaking at
close to $1/2$ for the binarized version when both networks have similar number
of edges. When employing the heuristic stopping criterion proposed in
Ref.~\cite{decelle_pseudolikelihood_2014}, we obtain a network with $E=24,196$
edges --- far denser that what we obtain with MDL. Due to these limitations, we
see no advantage of using decimation over our approach based on MDL.

\section{Empirical case study: network of microbial interactions}\label{sec:microbial}
\begin{figure*}
  \setlength{\tabcolsep}{0pt}
  \begin{tabular}{cccc}
    \multicolumn{2}{c}{(a) Human microbiome (HMP)} &
    \multicolumn{2}{c}{(b) Earth microbiome (EMP)} \\
    \multicolumn{2}{c}{\begin{overpic}[width=.5\textwidth]{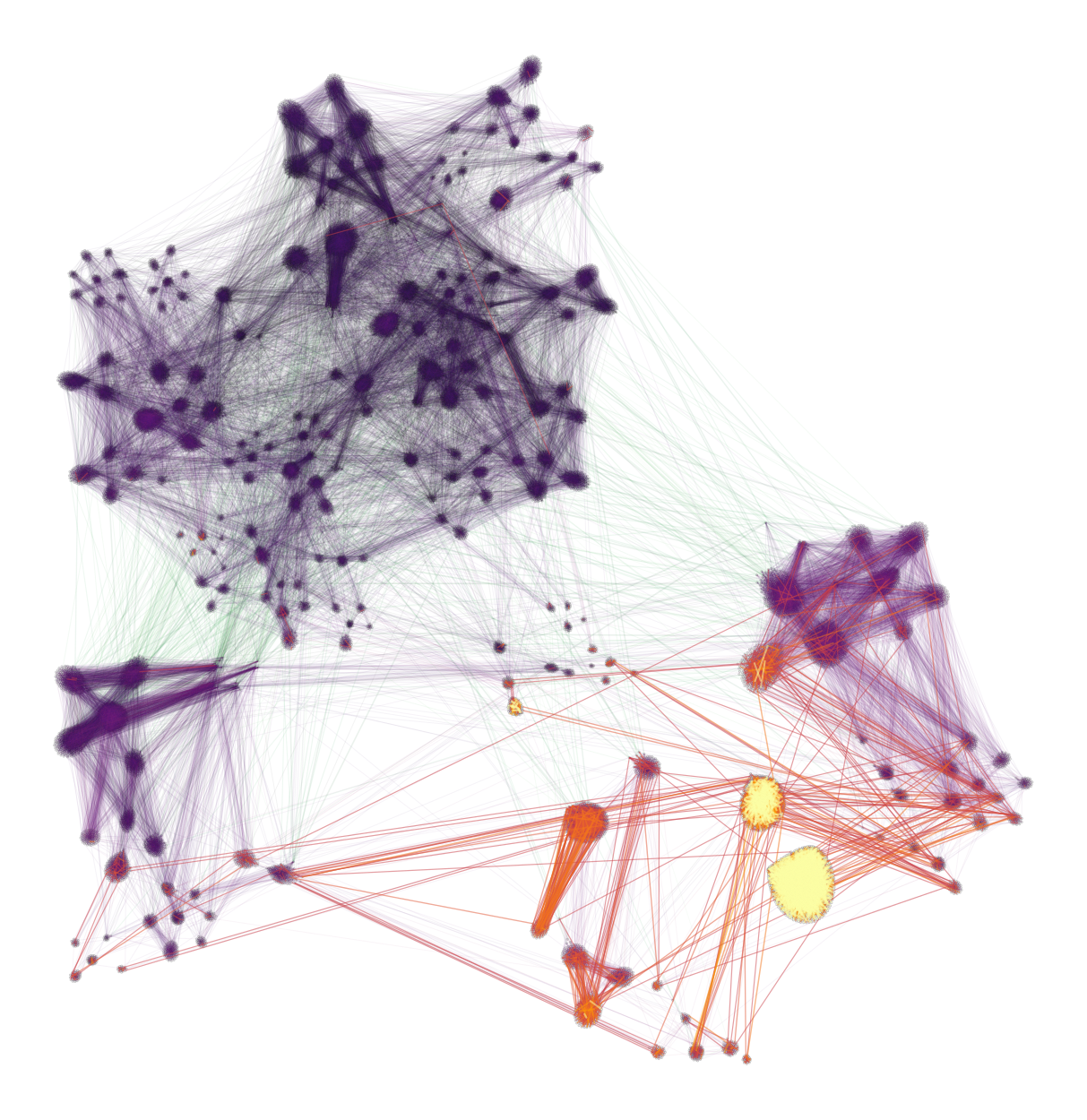}\end{overpic}} &
    \multicolumn{2}{c}{{\includegraphics[width=.5\textwidth,trim=0 -1.5cm 0 0]{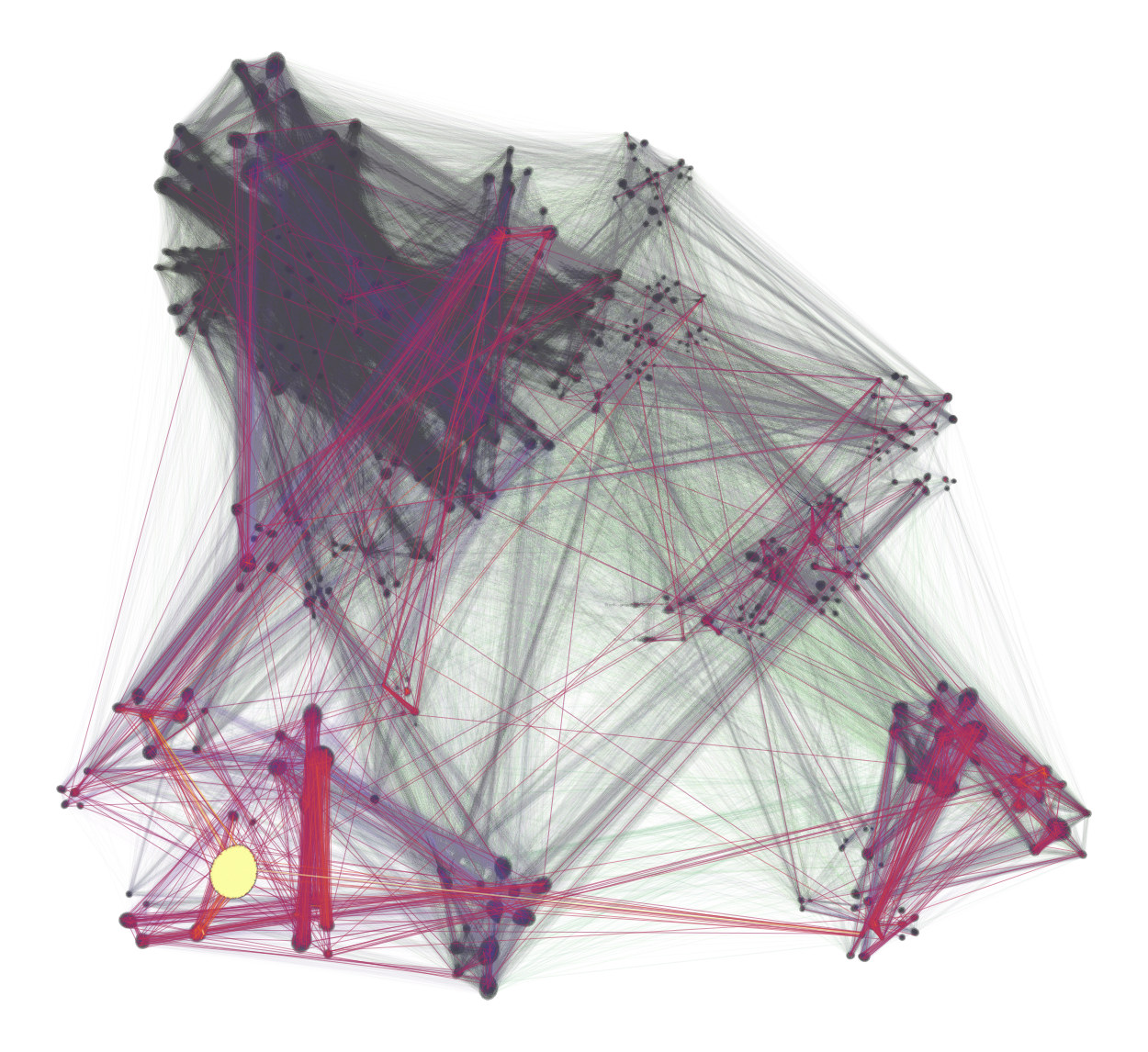}}}\\
    \multicolumn{2}{c}{
    \resizebox{.5\textwidth}{!}{\begin{tabular}{cccccc}
    \includegraphics[width=.166\textwidth]{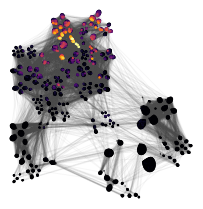} &
    \includegraphics[width=.166\textwidth]{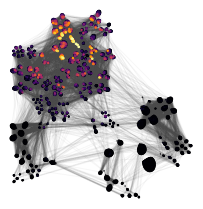} &
    \includegraphics[width=.166\textwidth]{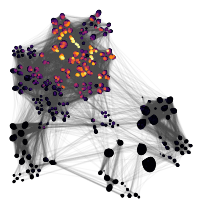} &
    \includegraphics[width=.166\textwidth]{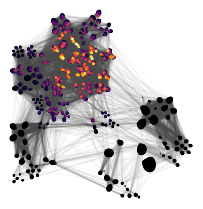} &
    \includegraphics[width=.166\textwidth]{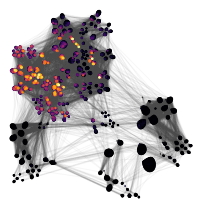} &
    \includegraphics[width=.166\textwidth]{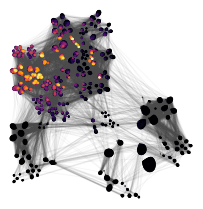} \\
    Attached keratinized gingiva & Buccal mucosa & Hard palate & Palatine tonsils & Subgingival plaque & Supragingival plaque \\
    \includegraphics[width=.166\textwidth]{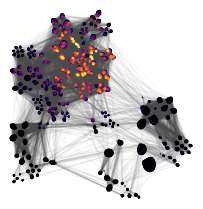} &
    \includegraphics[width=.166\textwidth]{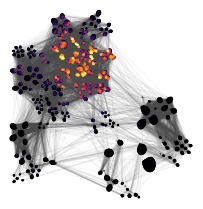} &
    \includegraphics[width=.166\textwidth]{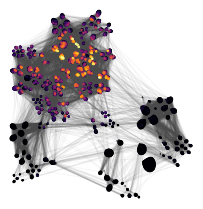} &
    \includegraphics[width=.166\textwidth]{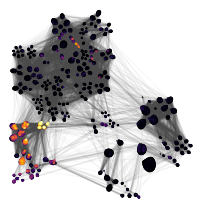} &
    \includegraphics[width=.166\textwidth]{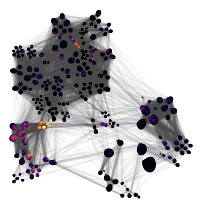} &
    \includegraphics[width=.166\textwidth]{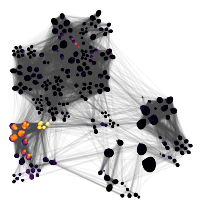} \\
    Throat & Tongue dorsum & Saliva & Anterior nares & Left antecubital fossa & Left retroauricular crease \\
    \includegraphics[width=.166\textwidth]{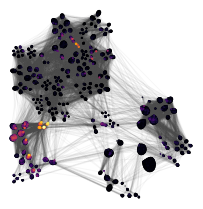} &
    \includegraphics[width=.166\textwidth]{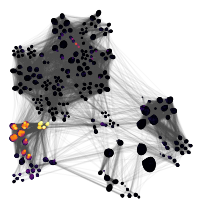} &
    \includegraphics[width=.166\textwidth]{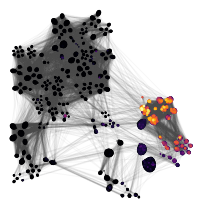} &
    \includegraphics[width=.166\textwidth]{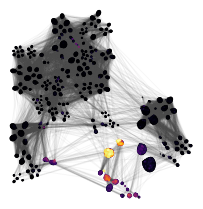} &
    \includegraphics[width=.166\textwidth]{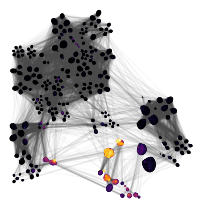} &
    \includegraphics[width=.166\textwidth]{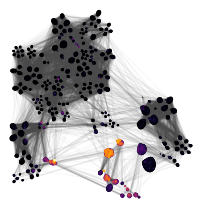} \\
    Right antecubital fossa & Right retroauricular crease & Stool & Posterior fornix & Mid-vagina & Vaginal introitus
    \end{tabular}}
    }&
    \multicolumn{2}{c}{
    \resizebox{.5\textwidth}{!}{\begin{tabular}{cccccc}
    \includegraphics[width=.166\textwidth]{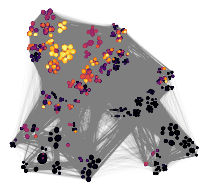} &
    \includegraphics[width=.166\textwidth]{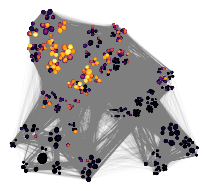} &
    \includegraphics[width=.166\textwidth]{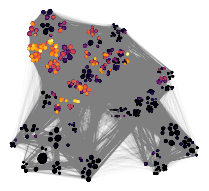} &
    \includegraphics[width=.166\textwidth]{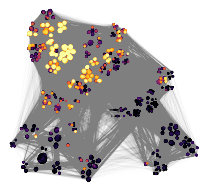} &
    \includegraphics[width=.166\textwidth]{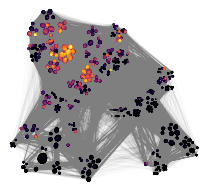} &
    \includegraphics[width=.166\textwidth]{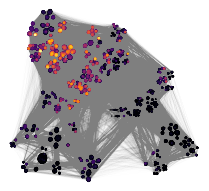}\\
    Cropland & Forest & Freshwater & Grassland & Tropical shrubland & Tundra \\
    \includegraphics[width=.166\textwidth]{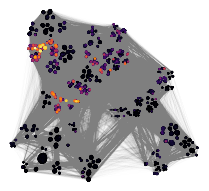} &
    \includegraphics[width=.166\textwidth]{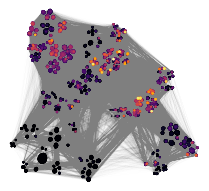} &
    \includegraphics[width=.166\textwidth]{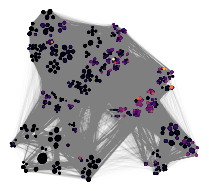} &
    \includegraphics[width=.166\textwidth]{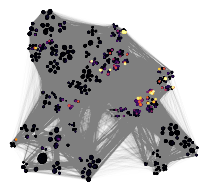}&
    \includegraphics[width=.166\textwidth]{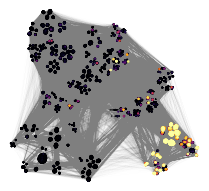} &
    \includegraphics[width=.166\textwidth]{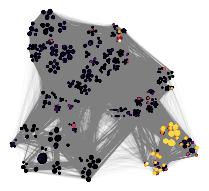} \\
    Large lake & Urban & Village & Coral reef & Dense settlement & Rangeland \\
    \includegraphics[width=.166\textwidth]{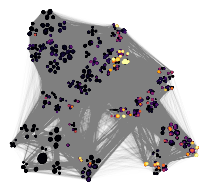} &
    \includegraphics[width=.166\textwidth]{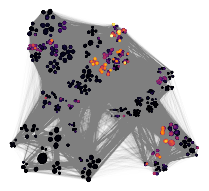}&
    \includegraphics[width=.166\textwidth]{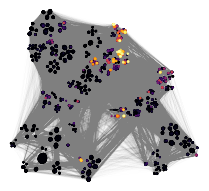}&
    \includegraphics[width=.166\textwidth]{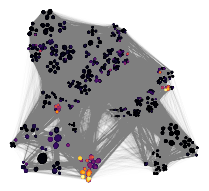} &
    \includegraphics[width=.166\textwidth]{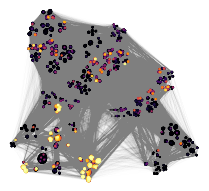} &
    \includegraphics[width=.166\textwidth]{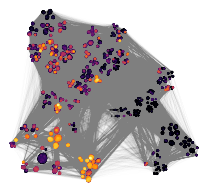} \\
    Mixed forest & Large river & Mediterranean shrubland & Marine & Marine pelagic & Marine benthic
    \end{tabular}}
    }\\\\
    \includegraphics[width=.25\textwidth]{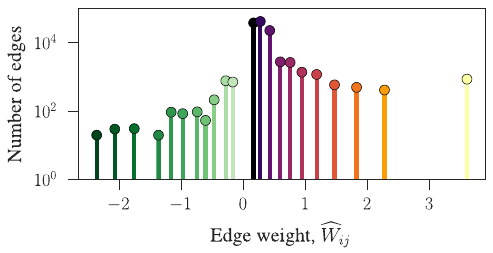} &
    \includegraphics[width=.25\textwidth]{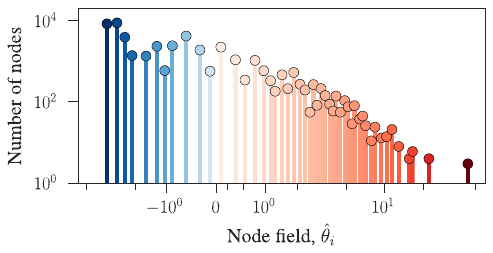} &
    \includegraphics[width=.24\textwidth]{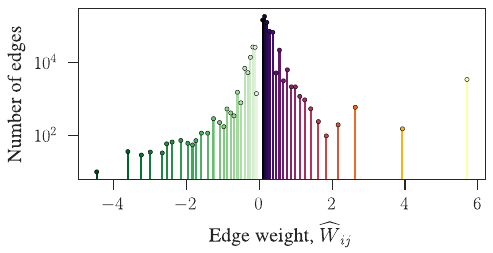} &
    \includegraphics[width=.24\textwidth]{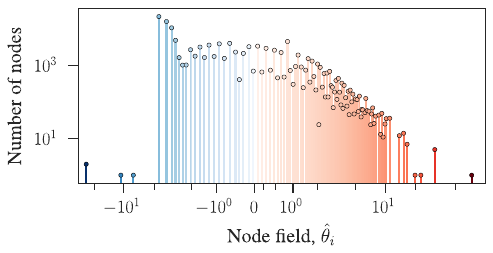} \\
    \multicolumn{2}{c}{\includegraphics[width=.48\textwidth]{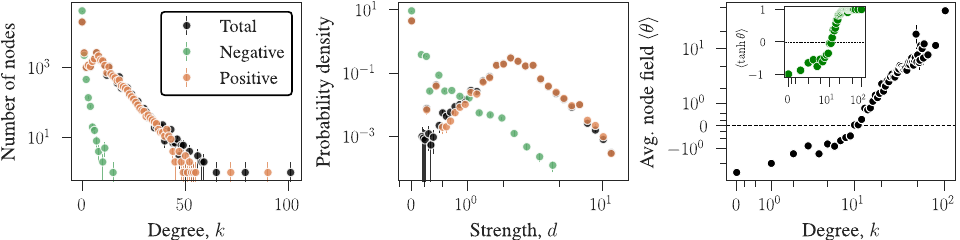}} &
    \multicolumn{2}{c}{\includegraphics[width=.48\textwidth]{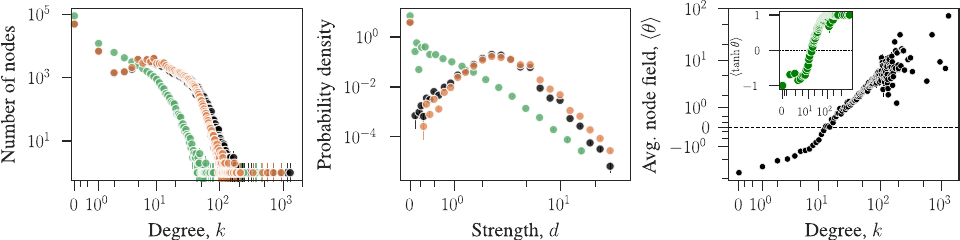}}
  \end{tabular}
  \caption{Reconstructed networks for the (a) human microbiome project (HMP)
    with $N=45,383$, $M=4,788$, and $E=122,804$ and (b) earth microbiome project
    (EMP) with $N=126,730$, $M=23,323$, and $E=735,868$, using the MDL
    regularization method, together with a SBM prior. The top panels show the
    networks obtained, with edge weights indicated as colors. The middle panel
    shows the OTU counts in (a) different body sites and (b) earth habitats (the
    color code is the same as the ``total count'' panels in
    Fig.~\ref{fig:taxonomy_emp}.). The bottom panel shows the edge weight
    distributions (with the same colors as the top panel), the node field
    distributions, the degree distributions (where node $i$ has total degree
    $k_{i}=\sum_{j}A_{ij}$, as well as positive and negative degrees,
    $k^{+}=\sum_{j}A_{ij}\mathds{1}_{W_{ij}>0}$ and
    $k^{-}=\sum_{j}A_{ij}\mathds{1}_{W_{ij}<0}$, respectively), the node
    strength distributions (where node $i$ has total strength
    $d_{i}=\sum_{j}W_{ij}$, as well as positive and negative strenght,
    $d_{i}^{+}=\sum_{j}W_{ij}\mathds{1}_{W_{ij}>0}$ and
    $d_{i}^{-}=\sum_{j}W_{ij}\mathds{1}_{W_{ij}<0}$, respectively), and the
    average node fields as a function of the degrees (with the expected
    ``magnetization'' in the inset). \label{fig:microbiome}}
\end{figure*}

\begin{figure}
  \setlength{\tabcolsep}{0pt}
  \begin{tabular}{cc}
    \begin{overpic}[width=.5\columnwidth]{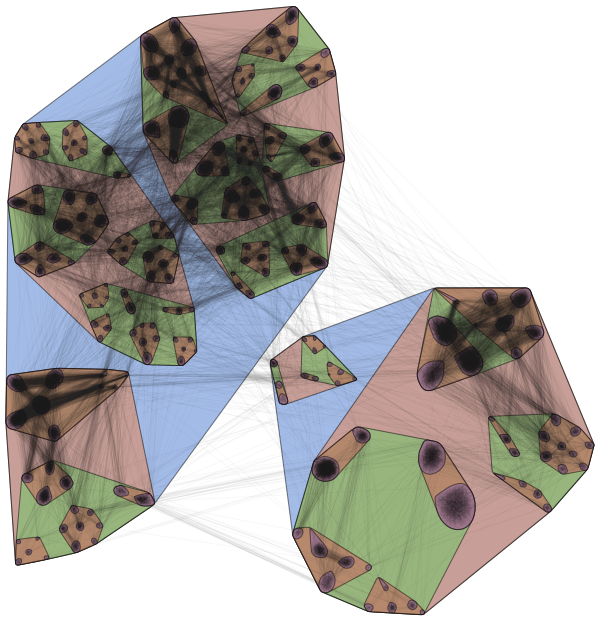}\put(0,0){(a)}\end{overpic} &
    \begin{overpic}[width=.5\columnwidth]{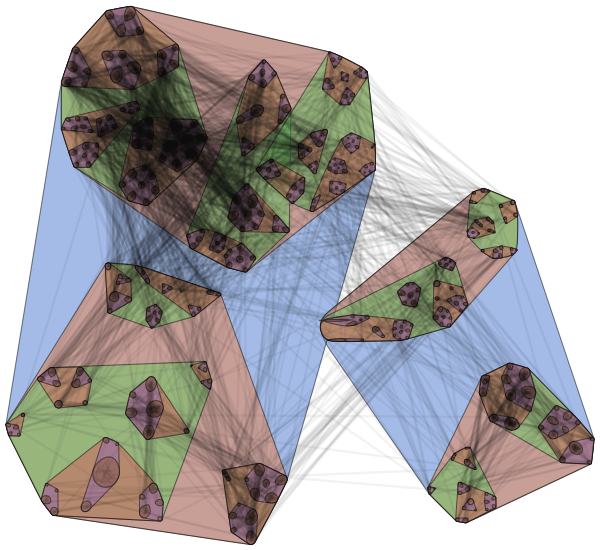}\put(0,0){(b)}\end{overpic}
  \end{tabular}
  \caption{Hierarchical community structure obtained with the nested
    SBM~\cite{peixoto_hierarchical_2014,peixoto_nonparametric_2017} integrated
    into the MDL regularization, for (a) the human microbiome project (HMP), and
    (b) the earth microbiome project (EMP), corresponding to the same networks
    shown in Fig.~\ref{fig:microbiome}. Each filled polygon represents the
    convex hull of the nodes that belong to the same group. Nested polygons
    represet nested hierarchical levels. \label{fig:nested-partition}}
\end{figure}

In this section we employ our regularization method for the analysis of
microbial
interactions~\cite{kurtz_sparse_2015,connor_using_2017,layeghifard_disentangling_2017,guseva_diversity_2022}.
We consider data from the human microbiome project
(HMP)~\cite{lloyd-price_strains_2017} and the earth microbiome project
(EMP)~\cite{gilbert_earth_2010}, which assemble thousands of samples of
microbial species abundances, collected, respectively, in various sites in the
human body and in diverse habitats across the globe. More precisely, these
datasets contain abundances derived from DNA sequencing, aggregated as
operational taxonomic units (OTUs)---groups of individuals closely related
according to their gene sequence, serving as a pragmatic proxy for individual
species. The HMP dataset contains $M=4,788$ samples involving $N=45,383$ OTUs,
and EMP contains $M=23,323$ samples involving $N=126,730$ OTUs [we used the
``Silva (CR)'' variant containing all samples].

The objective of our analysis is to obtain the underlying network of
interactions between species from their co-occurrence in samples. To account for
issues related with compositionality, i.e. the fact that the gene sequencing
techniques used in these studies can only provide the relative abundance between
species, instead of their absolute number, we will discard the magnitude of the
read counts, and consider only the presence or absence of species, i.e. if more
than zero counts have been observed for a given OTU, resulting in binary
observations. We will also not attempt to model the dependence with
environmental conditions and focus instead only on the mutual iterations between
species. These are both substantial omissions, but our purpose is not to fully
saturate the modelling requirements of this problem, but instead to demonstrate
how our regularization approach can be coupled with a particular generative
model to tackle a network reconstruction problem of this nature and magnitude
(to the best of our knowledge, with the exception of
Ref.~\cite{peixoto_scalable_2024}, the reconstruction based on the full HMP and
EMP datasets have not been performed so far).

We will model a given sample $\x\in \{-1,1\}^{N}$, where $x_{i}=-1$ if the
species $i$ is absent in the sample, and $x_{i}=1$ if it is present, according
to the Ising model, i.e. with probability
\begin{equation}
  P(\bm x | \bm W, \bm \theta) = \frac{\ee^{\sum_{i<j}W_{ij}x_{i}x_{j} + \sum_{i}\theta_{i}x_{i}}}{Z(\W, \bm \theta)},
\end{equation}
where $W_{ij}\in \mathbb{R}$ is the coupling strength between species $i$ and
$j$, $\theta_{i}$ determines the individual propensity of species $i$ to occur
in samples, independently of the other species, and $Z(\W, \bm \theta)$ is a
normalization constant. A value of $W_{ij}=0$ indicates conditional independence
between species, i.e. the presence or absence of species $i$ does not affect the
occurrence probability of species $j$, and vice versa, when all other species
take the same value. A value $W_{ij}>0$ indicates cooperation, since conditioning
on the presence of species $i$ increases the probability of observing $j$ (and
vice versa), and a value $W_{ij}<0$ indicates antagonism, since it has the
opposite effect.

We are interested in the MAP estimation
\begin{align}
  \widehat{\W}, \widehat{\bm \theta} &= \underset{\W,\bm\theta}{\argmax}\;P(\W,\bm\theta | \X)\\
                                     &= \underset{\W,\bm\theta}{\argmax}\;\log P(\X|\W,\bm\theta) \nonumber\\
  &\qquad\qquad\qquad + \log P(\W) + \log P(\bm \theta),
\end{align}
using the MDL regularization for both $\W$ and $\bm \theta$ discussed
previously, including also the SBM prior. (We will not include a comparison with
$L_{1}$, since the cross-validation procedure it requires becomes
computationally prohibitive for problems of this magnitude.)

In Fig.~\ref{fig:microbiome} we show the results of the reconstruction for both
datasets. The inferred networks are strongly modular, composed of large node
groups, connected between themselves largely by negative interactions, and a
hierarchical subdivision into smaller groups (see also
Fig.~\ref{fig:nested-partition}), connected between theselves by predominately
positive weights. As the middle panel of Fig.~\ref{fig:microbiome} shows, the
large-scale modular structure is strongly associated with the different body
sites for the HMP network, and the different habitats for EMP. This association
makes it easier to interpret the negative edges between the larger groups: the
different regions represent distinct microbiota, composed of specialized
organisms that occupy specific ecological niches. Since we have not included
dependence on environmental conditions into our model, this mutual exclusion
gets encoded by negative couplings. We also observe that nodes with low degree
tend to occur more seldomly (negative $\theta_{i}$ values) whereas node with
larger degrees tend to occur more frequently (positive $\theta_{i}$ values).

Strikingly, as we show in Appendix~\ref{app:taxonomy}, the groups found with
the SBM that is part of our MDL regularization correlate significantly with the
known taxonomic divisions of the microbial species considered, where most of the
groups found contain predominately a single genus, although the same genus can
be split into many inferred groups. This indicates different scales of
similarity and dissimilarity between taxonomically proximal species when it
comes to their interaction patterns with other species.

\subsubsection{Predicting stability and outcome of interventions}

In this section we explore the fact that our reconstructed networks equip us
with more than just a structural representation of the interactions between
species, but it also provides a generative model that can be used to answer
questions that cannot be obtained directly from the data. One particular
question that is of biological significance is the extent to which an ecological
system is stable to perturbations, and what are the outcomes of particular
interventions. One important concept for ecology is that of ``keystone''
species, defined as those that have a disproportionally strong impact in its
environment relative to its
abundance~\cite{paine_note_1969,mills_keystone-species_1993}. In the context of
microbial systems, keystone species are often associated with the impact they
have once they are forcibly removed from the system, measured by the number of
species that disappear as a consequence~\cite{mills_keystone-species_1993}.

The removal of individual species can be investigated with the Ising model in
the following way. First we observe that since the network structure is modular,
and the weights $\W$ are heterogeneous, the probability landscape for the
configurations $\bm x$ in such cases is typically multimodal, composed of many
``islands'' of high probability configurations separated by ``valleys'' of low
probability configurations---a property known as broken replica symmetry. If we
consider this model to correspond to the equilibrium configuration of a
dynamical system, then these islands amount to metastable configurations from
which the system would eventually escape, but only after a very long time that
grows exponentially with the total number of nodes. In this situation we can
decompose the overall probably of a configuration $\bm x$ as an average of all
these discrete metastable states---which we will henceforth call
\emph{macrostates}---, i.e.
\begin{equation}
  P(\bm x | \W,\bm\theta) = \sum_{\alpha}P(\bm x |\alpha)P(\alpha),
\end{equation}
with
$P(\bm x |\alpha)=\mathds{1}_{\bm x \in \alpha}P(\bm x | \W,\bm\theta)/P(\alpha)$
being the configuration distribution when trapped in macrostate $\alpha$, and
$P(\alpha)=\sum_{\bm x}\mathds{1}_{\bm x \in \alpha}P(\bm x | \W,\bm\theta)$ is
the asymptotic probability that this macrostate is eventually visited, if we run
the underlying dynamics for an infinitely long time.

The individual macrostates $\alpha$ have an important biological meaning, since
they would correspond to the actual configurations that we would encounter in
practice, as the time it would take to transition from one macrostate to another
would not be experimentally observable.

\begin{figure}
  \setlength{\tabcolsep}{0pt}
  \begin{tabular}{cc}
    \multicolumn{2}{c}{\begin{overpic}[width=\columnwidth]{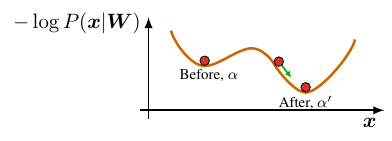}\put(0,5){(a)}\end{overpic}}\\
    Before, $\alpha$ & After, $\alpha'$ \\
    \framebox{\begin{overpic}[width=.47\columnwidth]{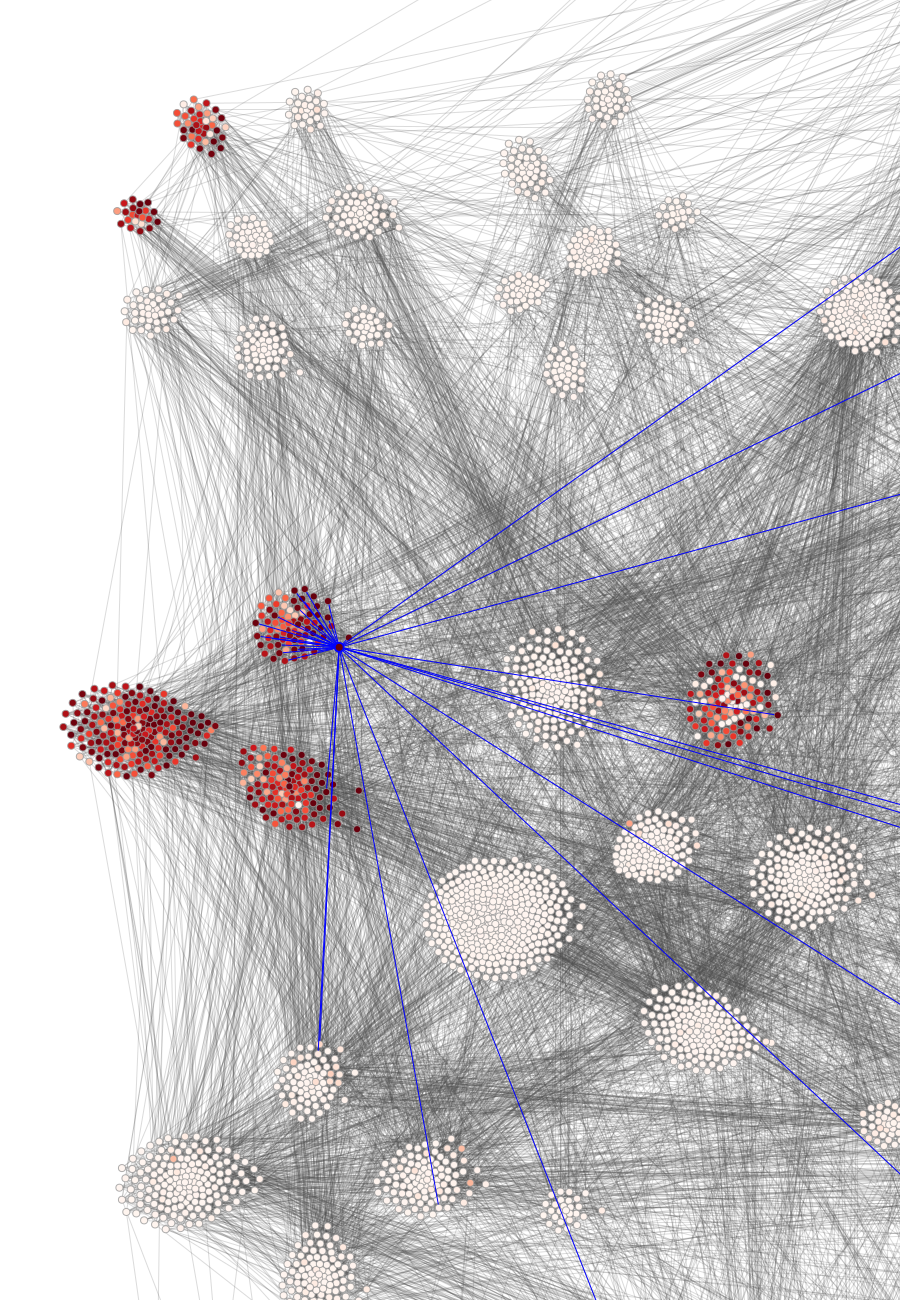}\put(0,0){(b)}\end{overpic}} &
    \framebox{\begin{overpic}[width=.47\columnwidth]{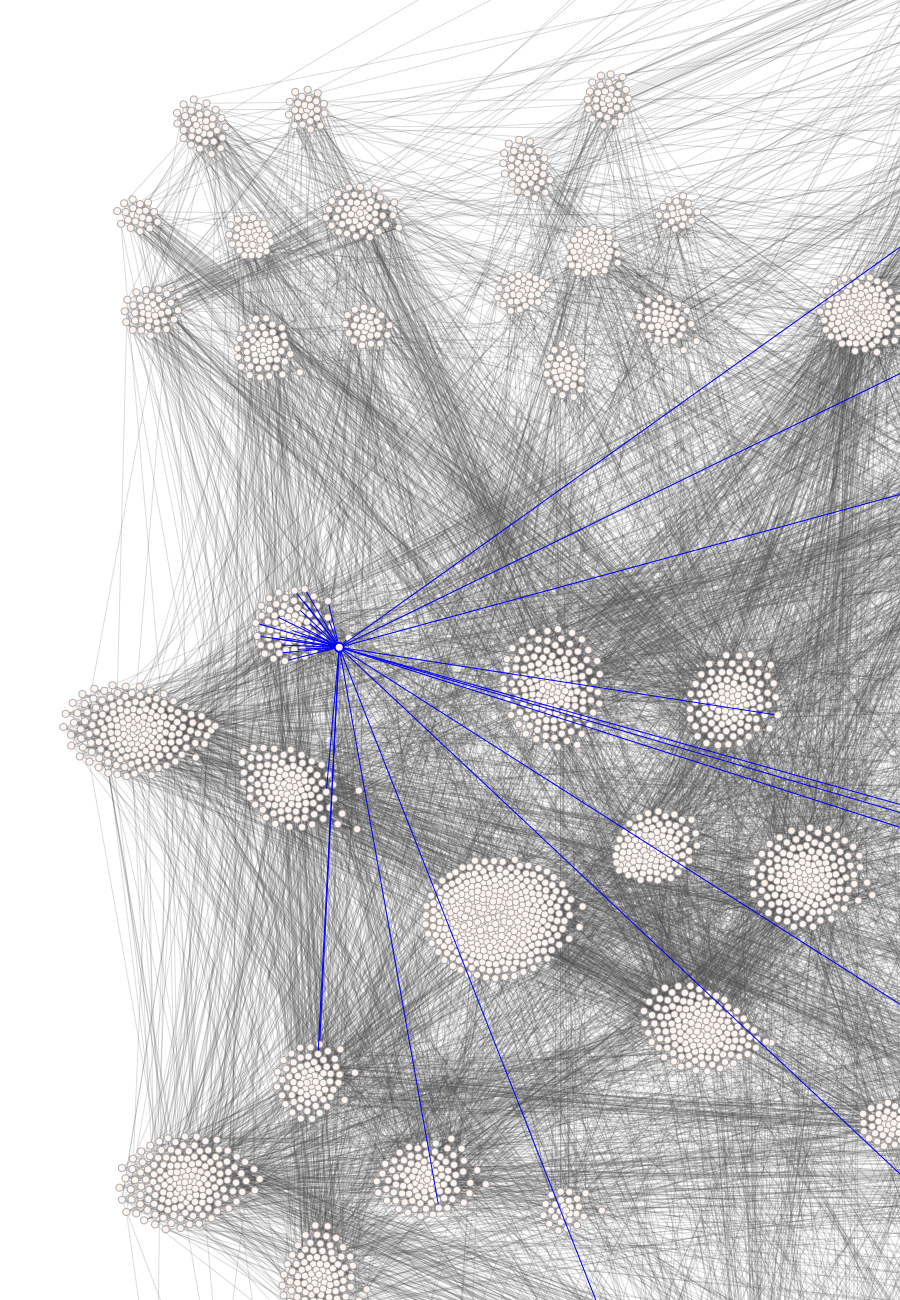}\put(0,0){(c)}\end{overpic}} \\
    \framebox{\begin{overpic}[width=.47\columnwidth]{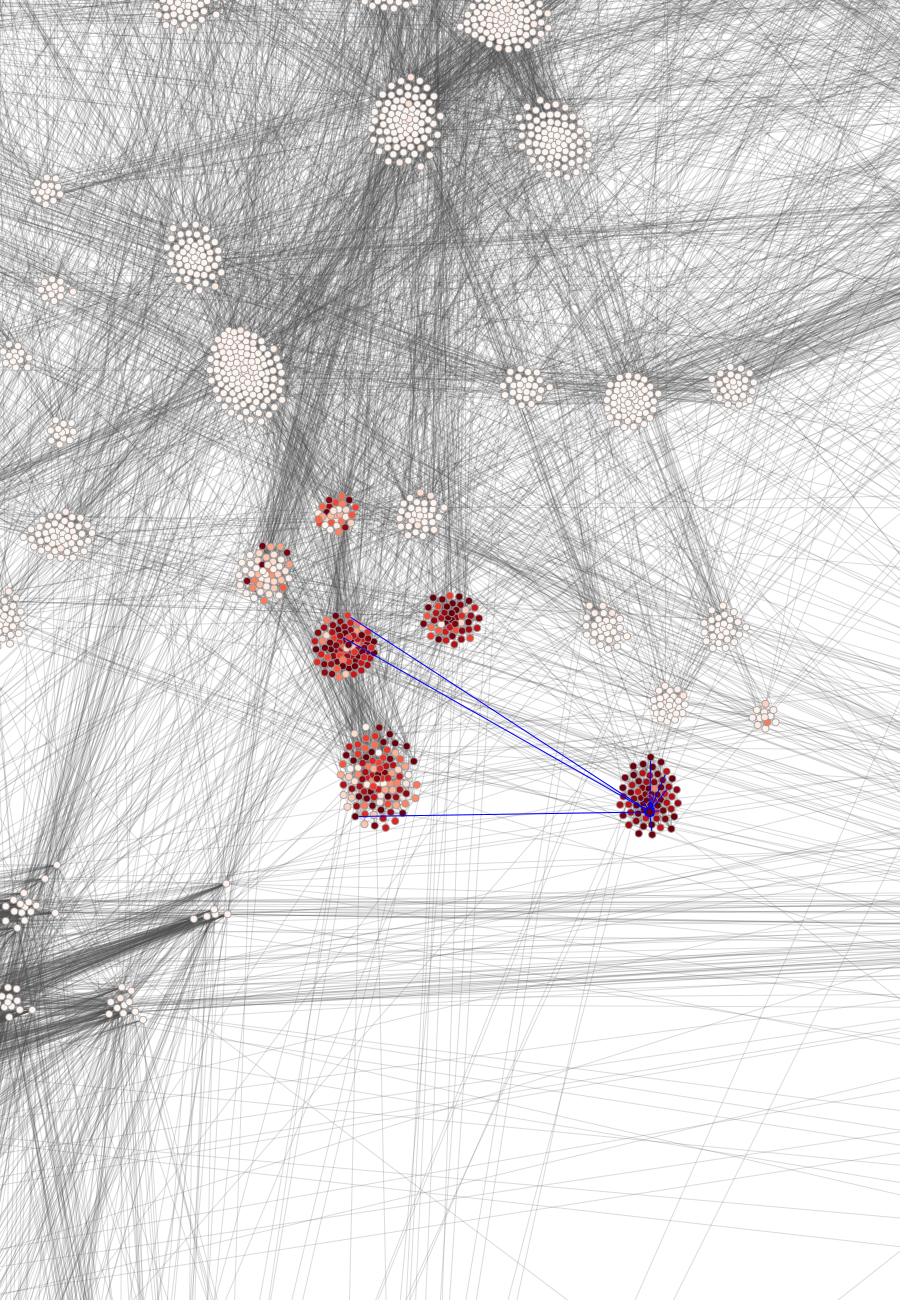}\put(0,0){\colorbox{white}{(d)}}\end{overpic}} &
    \framebox{\begin{overpic}[width=.47\columnwidth]{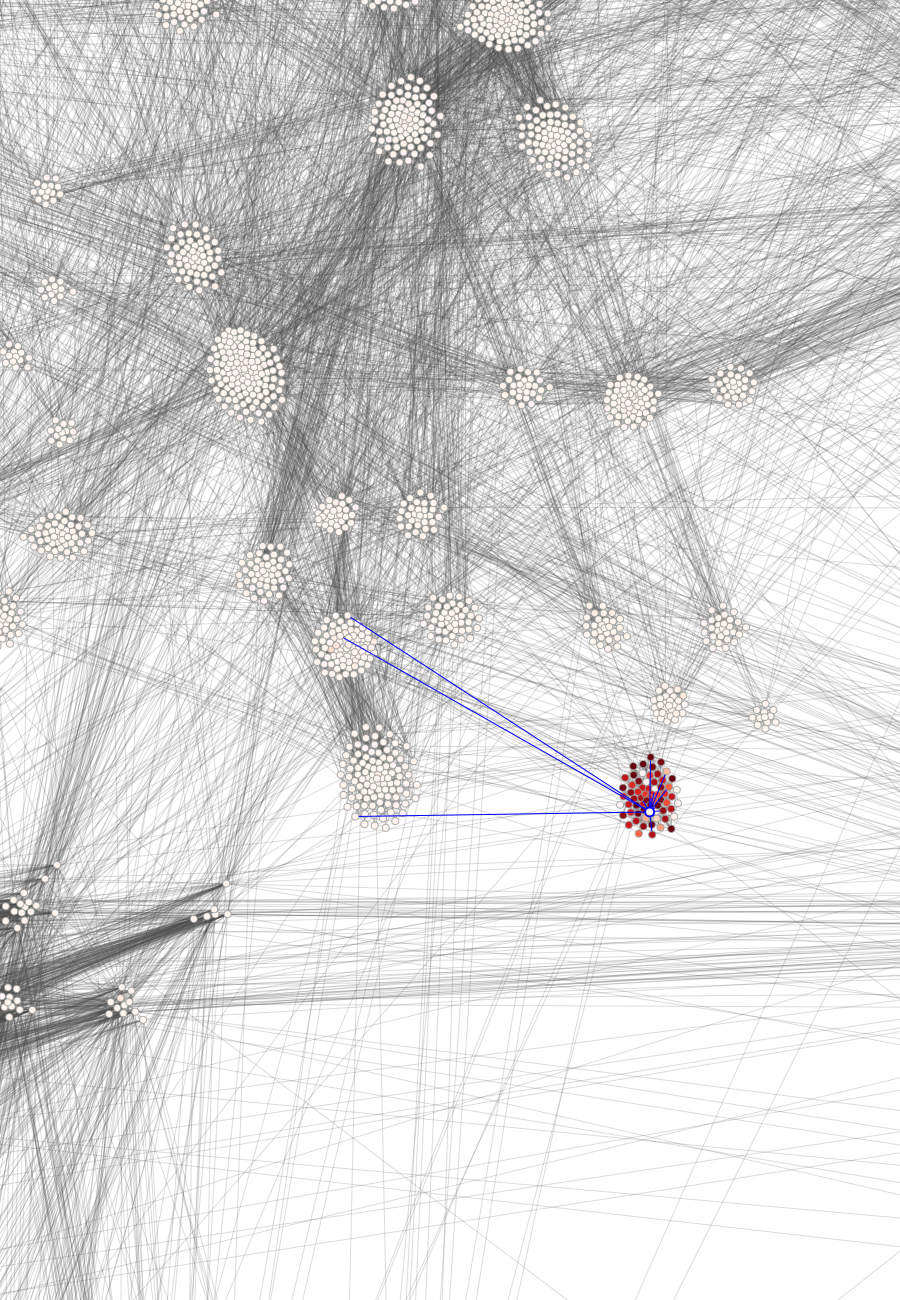}\put(0,0){\colorbox{white}{(e)}}\end{overpic}}
  \end{tabular}
  \caption{Examples of single-species perturbations that cause the system to
    move from a macrostate $\alpha$ to another $\alpha'$, as illustrated in
    panel (a), for the model inferred from the HMP data. Panels (b) and (d) show
    the macrostate before the perturbation, and (c) and (d) after. The node
    colors indicate the marginal expectations (white is absent and red is
    present), and the blue edges are the ones incident on the perturbed
    node.\label{fig:metastable}}
\end{figure}

We can then characterize the stability of the system precisely as its tendency
to escape from one such a macrostate after a small portion of the species is
perturbed (see Fig.~\ref{fig:metastable}a). In the case of a single species, we
can quantify this by comparing the marginal expectation of the presence of a
species $i$ in the unperturbed system, i.e.
\begin{equation}
  \bar{x}_{i}(\alpha) = \sum_{\bm x}x_{i}P(\bm x|\alpha),
\end{equation}
with the expectations obtained when another species $j\ne i$ is forced to take
value $x_{j}=-1$, i.e.
\begin{equation}
  \bar{x}'_{i}(\alpha,j) = \sum_{\bm x\setminus \{x_{j}\}}x_{i}P(\bm x \setminus \{x_{j}\}| x_{j}=-1,\W,\bm\theta,\alpha),
\end{equation}
where in the last equation the distribution should be interpreted as the
outcome of the dynamics if we initialize it with a configuration from
macrostate $\alpha$, allowing it to potentially escape from it.

We can then quantify the magnitude of the perturbation of node $j$ in macrostate
$\alpha$ via the expected number of additional species that disappear after it
is removed,
\begin{equation}
  z(j,\alpha) = \frac{1}{2}\sum_{i\neq j}\bar{x}_{i}(\alpha)-\bar{x}_{i}'(\alpha, j).
\end{equation}
If the perturbation is insufficient to dislodge the system to another macrostate
we will typically have $z(j,\alpha) \approx 0$, otherwise the magnitude of the
$z(j,\alpha)$ will indicate the difference between the macrostates. We could
then say that when the macrostate $\alpha$ is encountered, $j$
is a keystone species if $z(j,\alpha)$ is large.

\begin{figure}
  \includegraphics[width=\columnwidth]{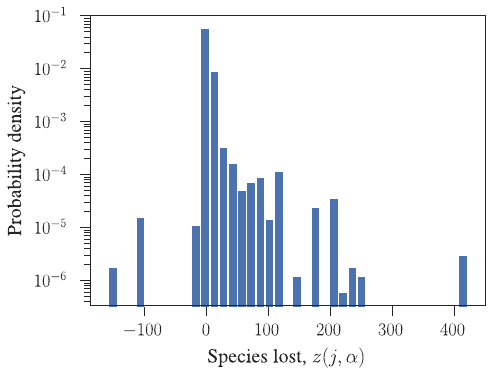}
  \caption{Probability distribution for the number of species lost $z(j,\alpha)$
    when a species $j$ chosen uniformly at random is removed from the system in
    macrostate $\alpha$, averaged over many macrostates, for the Ising model
    inferred from the HMP data.\label{fig:keystone}}
\end{figure}

In Fig.~\ref{fig:keystone} we show the distribution of $z(j,\alpha)$ for
macrostates simulated with the Ising model inferred from the HMP data, using the
belief-propagation (BP) method~\cite{mezard_information_2009}, initialized with
random messages. Although most perturbations are inconsequential, we observe
that sometimes hundreds of species can disappear (or even appear) after a single
species is removed.

In Fig.~\ref{fig:metastable} we highlight two perturbations with large
$z(j,\alpha)$ values, where in one case all existing species vanish, and another
where only a smaller subset remains. Importantly, the keystone species in these
cases do not directly influence all species lost, instead they are affected only
indirectly. In the corresponding dynamics, the move between macrostates would
occur through a cascade of extinctions, where the most immediate neighbors of
$j$ would first disappear, and then the neighbors' neighbors, an so on. Notably,
this kind of transition is typically irreversible, i.e. forcibly re-instating
the removed species will not result in a transition in the reverse direction to
the original macrostate. This kind of irreversible large-scale change that
results from a small perturbation characterizes a ``tipping point,'' whose
systematic prediction is major challenge in ecology and climate
science~\cite{dakos_ecosystem_2019}.

These predicted outcomes for the HMP data have only a tentative character,
since, as we mentioned before, we are omitting important aspects from the
model---such as the abundance magnitudes and environmental
factors~\cite{vinciotti_random_2023}---and making a strong modeling assumption
with the Ising model. Nevertheless, this example already illustrates how even an
elementary modelling assumption is sufficient to give us access to functional
properties of a latent system, inferred from empirical data, which comes as a
result of nonlinear emergent behavior mediated by the interactions between the
species. When complemented with more realistic generative models, this kind of
approach could be used to predict empirical outcomes, and be validated with
experiments or additional observational data. Towards this goal, appropriate
regularization is crucial, since a model that overfits will deliver neither
meaningful interpretations nor accurate predictions.

\section{Conclusion}\label{sec:conclusion}

We have presented a principled nonparametric regularization scheme based on
weight quantization and the MDL principle that can be used to perform network
reconstruction via statistical inference in a manner that adapts to the
statistical evidence available in the data. Our approach inherently produces the
simplest (i.e. sparsest) network whenever the data cannot justify a denser one,
and hence does not require the most appropriate number of edges to be known
beforehand, and instead produces this central information as an output, as is
often desired. Our approach does not rely on any specific property of the
generative model being used, other than it being conditioned on a weighted
adjacency matrix, and therefore it is applicable for a broad range of
reconstruction problems.

Unlike $L_{1}$ regularization, our method does not rely on weight shrinkage, and
therefore introduces less bias in the reconstruction, including a much reduced
tendency to infer spurious edges. Since it does not rely on cross-validation, it
requires only a single fit to the data, instead of repeated inferences as in the
case of $L_{1}$.

When combined with the subquadratic algorithm of
Ref.~\cite{peixoto_scalable_2024}, based on a stochastic second-neighbor search,
our regularization scheme can be used to reconstruct networks with a large
number of nodes, unlike most methods in current use, which have an algorithmic
complexity that is at best quadratic on the number of nodes, typically even
higher.

Our structured priors are extensible, and when coupled with inferential
community detection~\cite{peixoto_bayesian_2019} can also use latent modular
network structure to improve the regularization, and as a consequence also the
final reconstruction accuracy~\cite{peixoto_network_2019}.

As we have demonstrated with an empirical case study on microbial interactions,
our method is effective at uncovering functional latent large-scale complex
systems from empirical data, which can then be used to make predictions about
future behavior, the effect of interventions, and the presence of tipping
points.

\begin{acknowledgments}
This work has been funded by the Vienna Science and Technology Fund (WWTF) and
by the State of Lower Austria [Grant ID: 10.47379/ESS22032]. It is also in part
the result of research conducted for Central European University, Private
University. It was made possible by the CEU Open Access Fund.
\end{acknowledgments}

\bibliography{bib}

%\newpage
\appendix
\onecolumngrid
\section{Generative models}\label{app:models}

In our examples we use two generative models: the equilibrium Ising
model~\cite{nguyen_inverse_2017} and the kinetic Ising model. The equilibrium Ising
model is a distribution on $N$ binary variables $\bm{x} \in \{-1,1\}^{N}$ given
by
\begin{equation}\label{eq:isingboltz}
  P(\bm x | \bm W, \bm \theta) = \frac{\ee^{\sum_{i<j}W_{ij}x_{i}x_{j} + \sum_{i}\theta_{i}x_{i}}}{Z(\W, \bm \theta)},
\end{equation}
with $\theta_{i}$ being a local field on node $i$, and
$Z(\W,\bm\theta)=\sum_{\bm x}\ee^{\sum_{i<j}W_{ij}x_{i}x_{j} + \sum_{i}\theta_{i}x_{i}}$
a normalization constant. Since this normalization cannot be computed
analytically in closed form, we make use of the pseudolikelihood
approximation~\cite{besag_spatial_1974},
\begin{align}\label{eq:ising_pseudo}
  P(\bm x | \bm W, \bm \theta) &= \prod_{i}P(x_{i}|\bm x\setminus {x_{i}}, \bm W, \bm \theta)\\
  &= \prod_{i}\frac{\ee^{x_{i}(\sum_{j}W_{ij}x_{j} + \theta_i)}}{2\cosh(\sum_{j}W_{ij}x_{j} + \theta_i)},
\end{align}
as it gives asymptotically correct results and has excellent performance in
practice~\cite{mozeika_consistent_2014,nguyen_inverse_2017}.

The kinetic Ising model is a Markov chain with transition probabilities
conditioned on the same parameters as before, given  by
\begin{equation}\label{eq:ising_kinetic}
  P(\bm x(t+1) |\bm x(t), \bm W, \bm \theta) =
  \prod_{i}\frac{\ee^{x_{i}(t+1)(\sum_{j}W_{ij}x_{j}(t) + \theta_i)}}{2\cosh(\sum_{j}W_{ij}x_{j}(t) + \theta_i)}.
\end{equation}
In the case of the zero-valued Ising model with $\bm{x}\in \{-1,0,1\}^{N}$, but following the same Eq.~\ref{eq:isingboltz}, the normalization of Eqs.~\ref{eq:ising_pseudo} and~\ref{eq:ising_kinetic} change from $2\cosh(\cdot)$ to  $1+2\cosh(\cdot)$.

\section{Species taxonomies}\label{app:taxonomy}

\begin{figure*}[b!]
  \setlength{\tabcolsep}{1em}
  \smaller
  \resizebox{\textwidth}{!}{\begin{tabular}{ccc}
    \multicolumn{3}{c}{\larger[2]Human microbiome project (HMP)}\\[1em]
    Phylum & Class & Order \\[1.5em]
    \begin{overpic}[width=.33\textwidth]{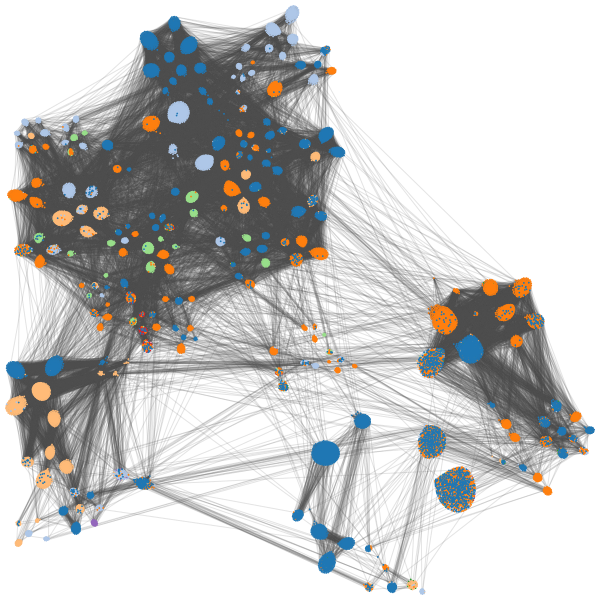}
      \put(70,55){\includegraphics[width=.09\textwidth]{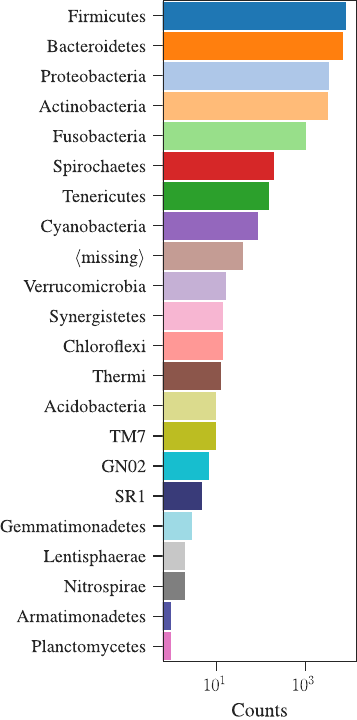}}
    \end{overpic} &
    \begin{overpic}[width=.33\textwidth]{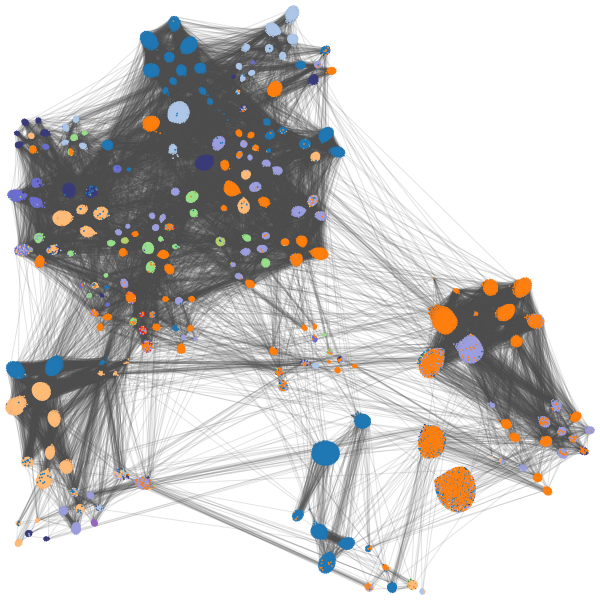}
      \put(70,55){\includegraphics[width=.09\textwidth]{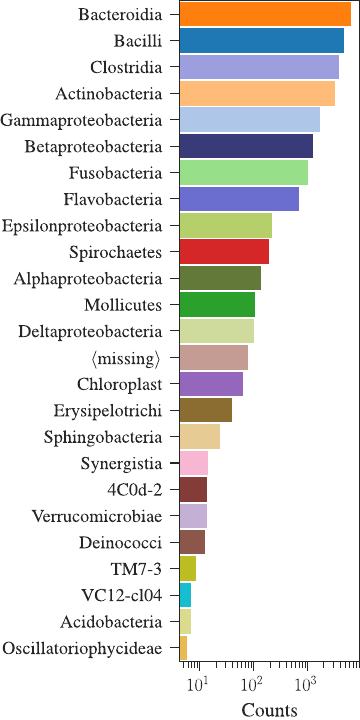}}
    \end{overpic} &
    \begin{overpic}[width=.33\textwidth]{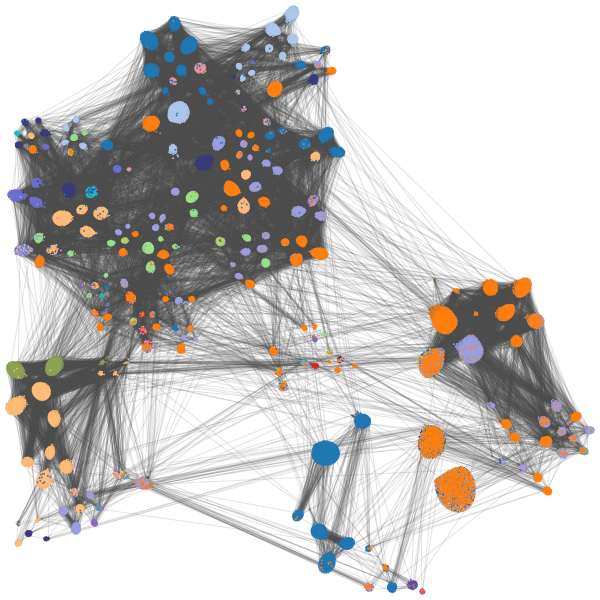}
      \put(70,55){\includegraphics[width=.09\textwidth]{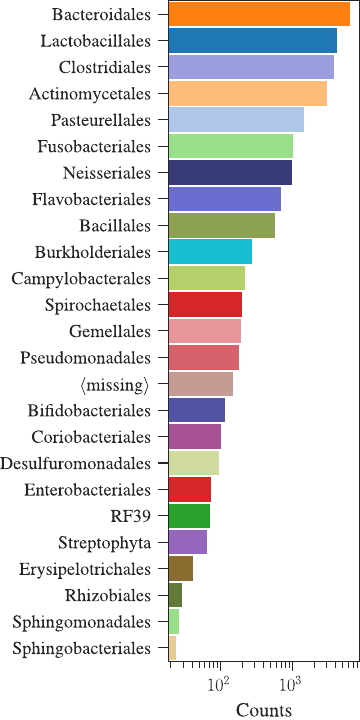}}
    \end{overpic} \\
    Family & Genus & Total counts \\
    \begin{overpic}[width=.33\textwidth]{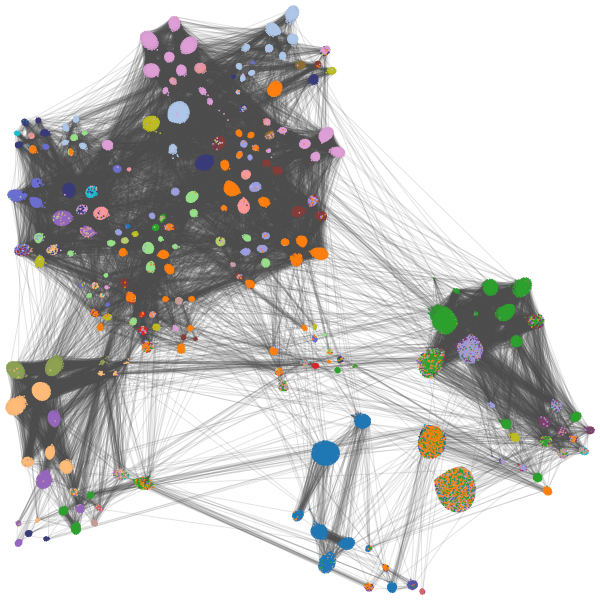}
      \put(70,55){\includegraphics[width=.09\textwidth]{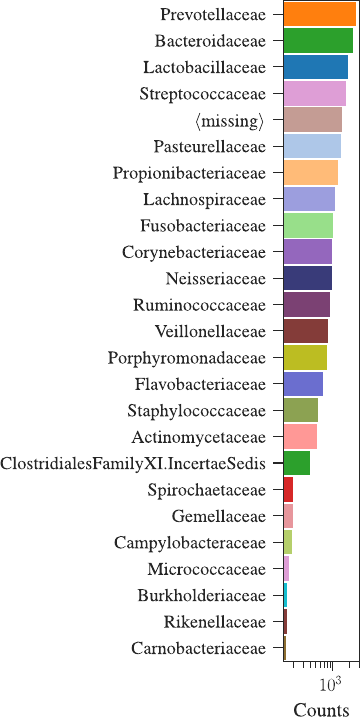}}
    \end{overpic} &
    \begin{overpic}[width=.33\textwidth]{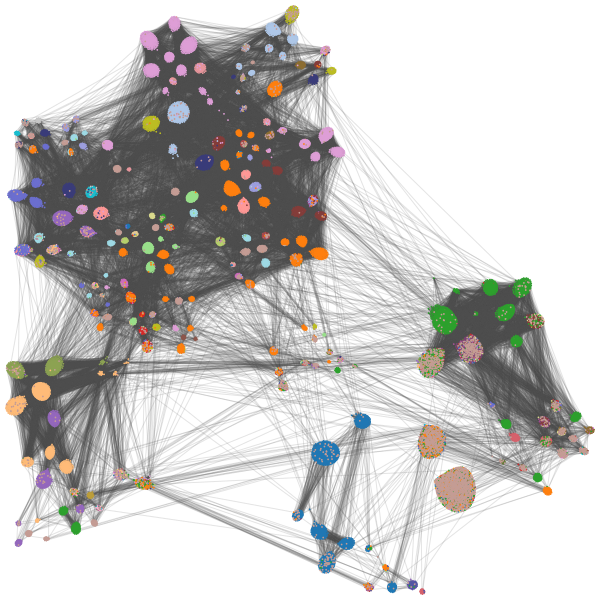}
      \put(70,55){\includegraphics[width=.09\textwidth]{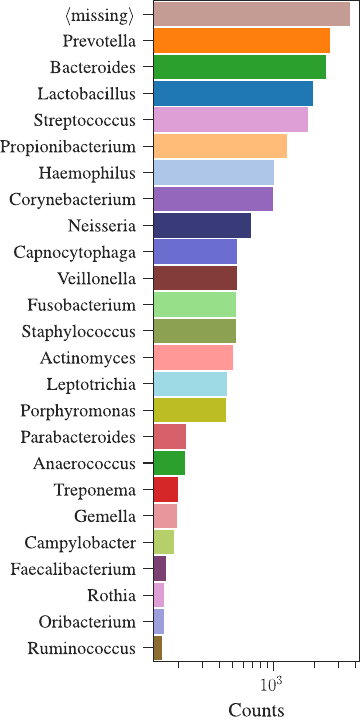}}
    \end{overpic} &
    \includegraphics[width=.33\textwidth]{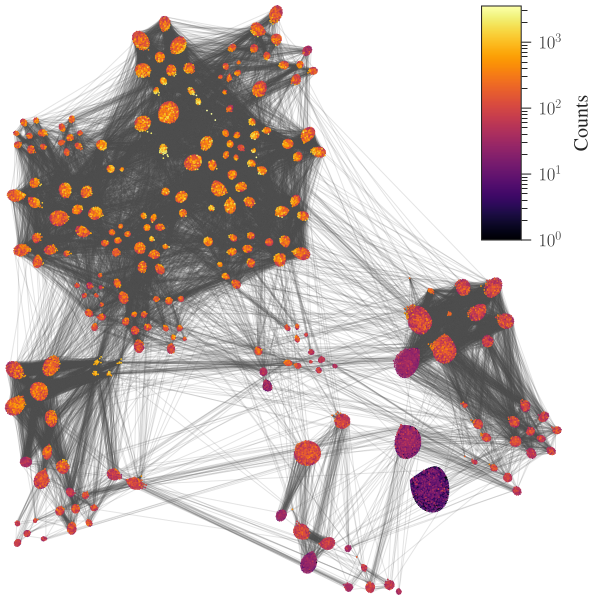}
  \end{tabular}}
% \caption{Taxonomic classification of species for the human microbiome project
%   (HMP), overlaid on the inferred networks of Fig.~\ref{fig:microbiome},
%   together with the total species count over all samples.\label{fig:taxonomy_hmp}}
% \end{figure*}

% \begin{figure*}[t!]
%   \setlength{\tabcolsep}{1em}
%   \smaller
  \resizebox{\textwidth}{!}{\begin{tabular}{ccc}
    \multicolumn{3}{c}{\larger[2]Earth microbiome project (EMP)}\\[1em]
    Phylum & Class & Order \\
    \begin{overpic}[width=.33\textwidth]{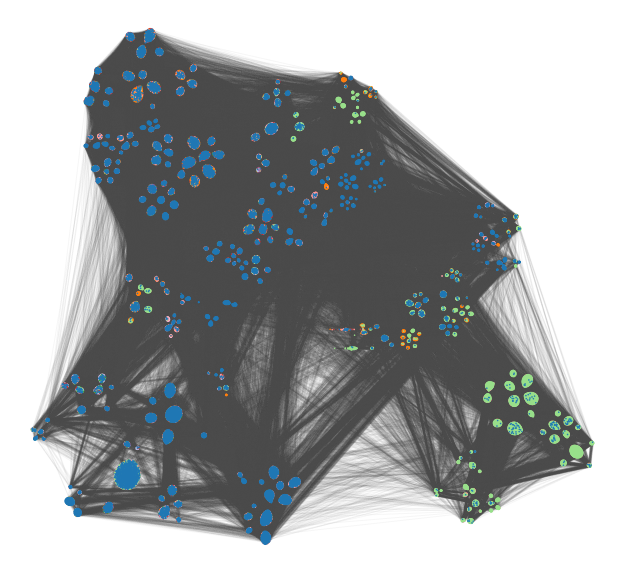}
      \put(86,30){\includegraphics[width=.09\textwidth]{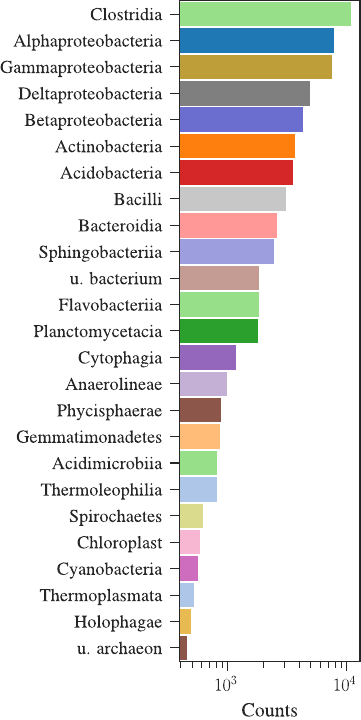}}
    \end{overpic} &
    \begin{overpic}[width=.33\textwidth]{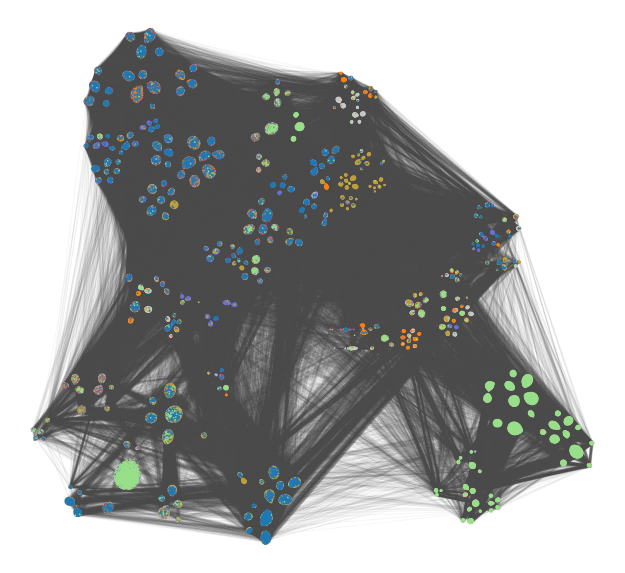}
      \put(86,30){\includegraphics[width=.09\textwidth]{emp-pa-ising-clineage-level2-leg.pdf}}
    \end{overpic} &
    \begin{overpic}[width=.33\textwidth]{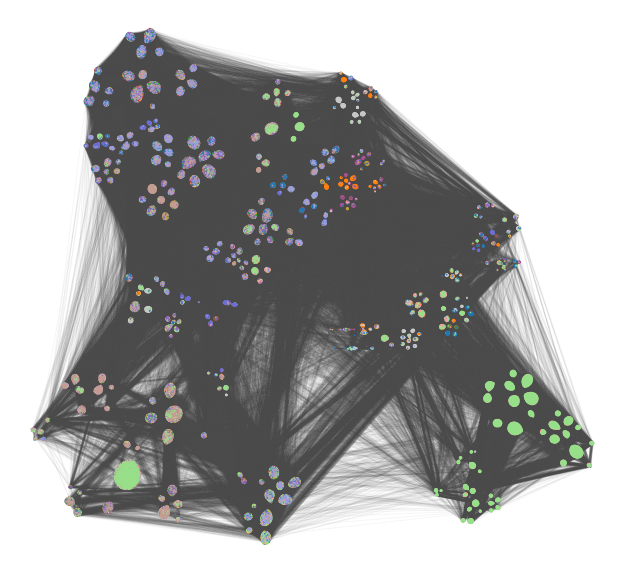}
      \put(86,30){\includegraphics[width=.09\textwidth]{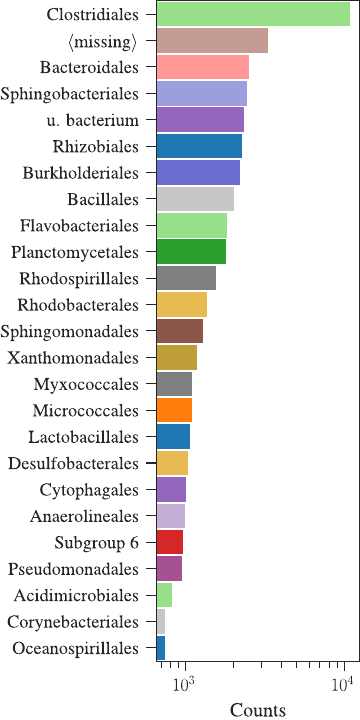}}
    \end{overpic} \\
    Family & Genus & Total counts \\
    \begin{overpic}[width=.33\textwidth]{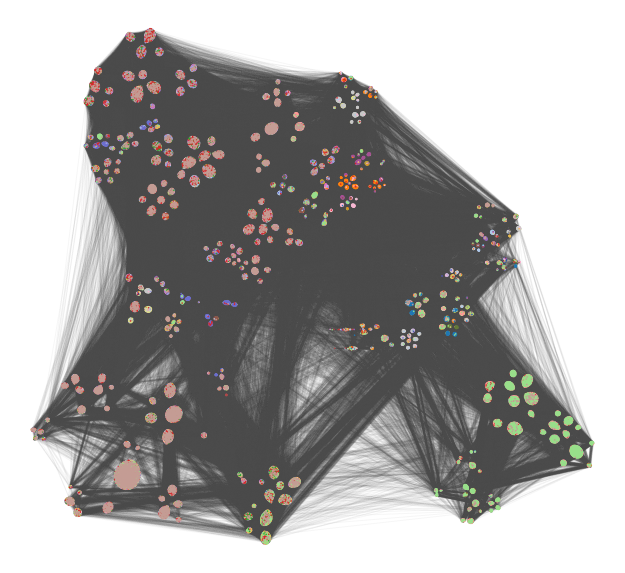}
      \put(86,30){\includegraphics[width=.09\textwidth]{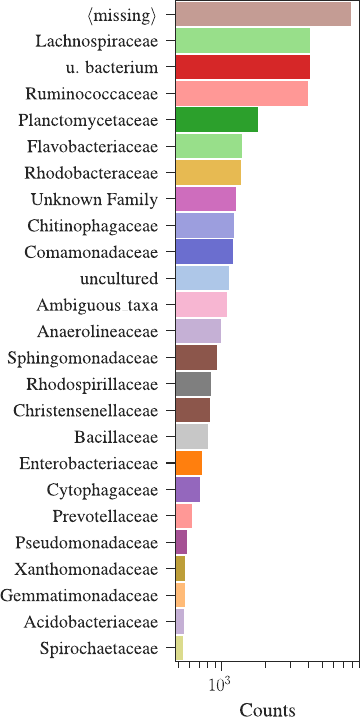}}
    \end{overpic} &
    \begin{overpic}[width=.33\textwidth]{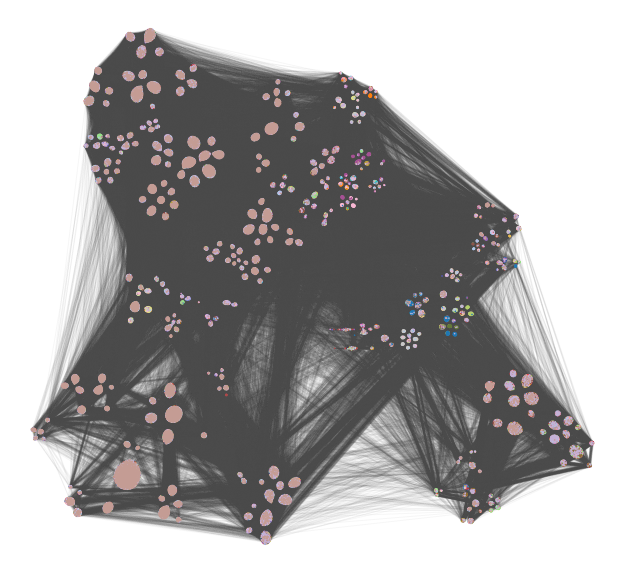}
      \put(86,30){\includegraphics[width=.09\textwidth]{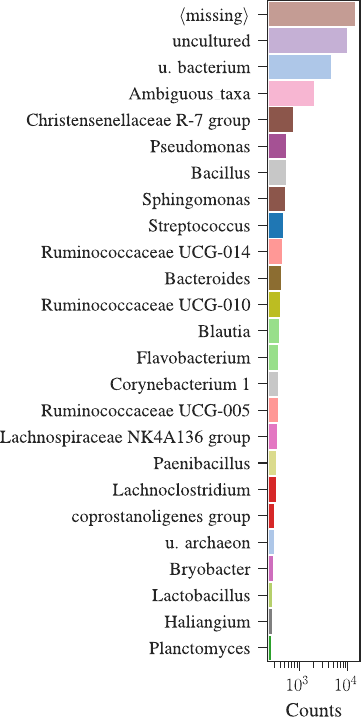}}
    \end{overpic} &
    \includegraphics[width=.33\textwidth]{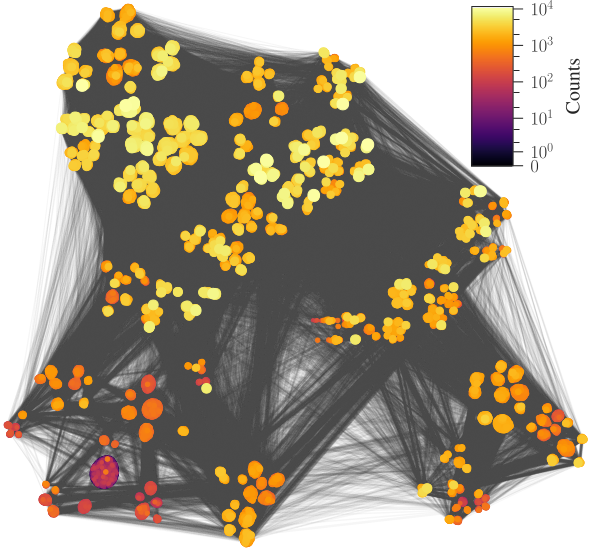}
  \end{tabular}}
\caption{Taxonomic classification of species for the human microbiome project
  (HMP) and the earth microbiome project (EMP), overlaid on the inferred networks of
  Fig.~\ref{fig:microbiome}, together with the total species count over all
  samples.\label{fig:taxonomy_emp}}
\end{figure*}

In Fig.~\ref{fig:taxonomy_emp} we show the taxonomic classification of the
species for the reconstructed networks of Fig.~\ref{fig:microbiome}. The
taxonomy consists of the phylum, class, order, family, and genus of each OTU.
Each taxonomic category is shown as node colors, as indicated in the legend. In
both cases we observe that most groups inferred with the SBM incorporated in our
regularization scheme are compatible with the taxonomic division, since they
tend to contain one dominating taxonomic category, at all levels --- although a
single category is often split into many groups, indicating structures that are
not captured solely by taxonomy.

\end{document}